\documentclass{article} % For LaTeX2e
\usepackage[usenames,dvipsnames]{xcolor}
\usepackage{iclr2023_conference,times}

\usepackage[utf8]{inputenc} % allow utf-8 input
\usepackage[T1]{fontenc}    % use 8-bit T1 fonts
\usepackage{hyperref}       % hyperlinks
\usepackage{url}            % simple URL typesetting
\usepackage{booktabs}       % professional-quality tables
\usepackage{amsfonts}       % blackboard math symbols
\usepackage{nicefrac}       % compact symbols for 1/2, etc.
\usepackage{microtype}      % microtypography
\usepackage{amssymb,amsmath,dsfont,bm}
\usepackage{enumitem}
\usepackage{tabularx}
\usepackage{tablefootnote}
\usepackage[normalem]{ulem}
\usepackage{prftree}
\usepackage{tikz}
\usetikzlibrary{decorations.pathreplacing,calligraphy,positioning,shapes.multipart}
\usepackage[capbesideposition=outside]{floatrow}
\usepackage[ruled,linesnumbered]{algorithm2e}
\usepackage{soul}
\usepackage{colortbl}
\usepackage{hhline}
\usepackage{pifont}
\usepackage{setspace}

\newcommand{\hlc}[2][yellow]{{%
    \colorlet{foo}{#1}%
    \sethlcolor{foo}\hl{#2}}%
}

\usepackage{titlesec}
\titlespacing\section{0pt}{1.0ex plus 0.3ex minus 0.2ex}{0.5ex plus 0.3ex minus 0.2ex}
\titlespacing\subsection{0pt}{0.7ex plus 0.3ex minus 0.2ex}{0.5ex plus 0.3ex}
\titlespacing\subsubsection{0pt}{0.7ex plus 0.3ex minus 0.2ex}{0.4ex plus 0.3ex}
\titlespacing\paragraph{0pt}{-1.6pt}{1.6ex plus 0.1ex}

\makeatletter

\newcommand{\cmark}{\ding{51}}
\newcommand{\xmark}{\ding{55}}

\definecolor{pale_green}{rgb}{0.55,0.75,0.60}
\definecolor{pale_red}{rgb}{0.90,0.61,0.58}
\definecolor{pale_yellow}{rgb}{0.95,0.92,0.72}

\def\SOUL@hlpreamble{%
\setul{}{2.77ex}%         !!!change this value!!! default is 2.5ex
\let\SOUL@stcolor\SOUL@hlcolor
\SOUL@stpreamble
}

\renewcommand{\algocf@makecaption@ruled}[2]{%
\global\sbox\algocf@capbox{\hskip\AlCapHSkip % left indent of the box
\addtolength{\algocf@lcaptionbox}{-2\AlCapHSkip}% reduce box width to account for symmetric margins
\parbox[t]{\algocf@lcaptionbox}{\algocf@captiontext{#1}{#2}}}% print the caption
}%

\makeatother

\newcommand{\dataset}{\textsc{PrOntoQA}}

\usepackage{graphicx}
\usepackage{wrapfig}
\usepackage{caption}

\title{Language Models Are Greedy Reasoners: A Systematic Formal Analysis of Chain-of-Thought}

\author{Abulhair Saparov \& He He \\
Center for Data Science, New York University, New York, NY 10011, USA \\
\texttt{\{as17582,hhe\}@nyu.edu}
}

\iclrfinalcopy % Uncomment for camera-ready version, but NOT for submission.
\begin{document}

\maketitle

%\abucomment{TODO: cite natural logic in related work where we mention reasoning directly over natural language}
%\abucomment{TODO: cite Hanlin's paper}
%\abucomment{TODO: remove \textbackslash{}pagestyle\{plain\}}
%\abucomment{TODO: add examples of results showcasing a misleading step leading to an error, and clarify some of the parts Daniel Khashabi mentioned}

\begin{abstract}
%Reasoning with pretrained large language models (LLMs) is a nascent area that is growing rapidly with broad applications in question-answering (QA), natural language inference (NLI), and multimodal tasks. 
Large language models (LLMs) have shown remarkable reasoning capabilities given chain-of-thought prompts (examples with intermediate reasoning steps).
%However, the extent of the reasoning ability of LLMs is not known: Which rules of deduction did the model acquire from pretraining? Can the model apply these rules accurately, especially in compositional settings and in new unseen contexts?
Existing benchmarks measure reasoning ability indirectly, by evaluating accuracy on downstream tasks such as mathematical reasoning. %Many past attempts to gauge the reasoning ability of LLMs focused narrowly on mathematical reasoning, which may not provide a representative evaluation of reasoning ability more broadly.
%In addition, LLMs rely on spurious correlations and heuristics in many benchmarks, such as lexical overlap in QA and NLI benchmarks, and this has confounded efforts to measure their reasoning ability.
However, it is unclear \emph{how} these models obtain the answers and whether they rely on simple heuristics rather than the generated chain-of-thought.
To enable systematic exploration of the reasoning ability of LLMs,
we present a new synthetic question-answering dataset called \dataset, where
each example is generated from a synthetic world model represented in first-order logic. %The sentences are syntactically simple, in order to focus the evaluation on reasoning rather than parsing.
This allows us to parse the generated chain-of-thought into symbolic proofs %over first-logic logic expressions,
for formal analysis.
%Evaluating the proofs directly is not prone to heuristics, unlike evaluation of the predicted label. 
%We analyze the proofs predicted by
Our analysis on \textsc{InstructGPT} and \textsc{GPT-3} shows that LLMs are quite capable of making correct individual deduction steps, and so are generally capable of reasoning, even in fictional contexts.
%\hh{This sentence is too generic: can we say they produce valid proofs instead of relying on heuristics?}\abucomment{hmm, after they make a misleading step, they will usually continue making valid steps and eventually make an invalid step to try to return to the conclusion of the proof. maybe we could just say they're ``quite capable of producing valid proof steps?''}.
However, they have difficulty with \emph{proof planning}: When %the model encounters a point in the proof where
multiple valid deduction steps are available, %it sometimes takes the wrong step, and this often leads the model to fail to complete the proof and produce incorrect output. 
they are not able to systematically explore the different options.
%Finally, we suggest ways to address these limitations in future work.
\end{abstract}

\section{Introduction}

The ability to reason---drawing new conclusions from provided facts---is a hallmark of human intelligence.
Recently, \emph{chain-of-thought} (CoT) prompting has enabled large language models (LLMs) to perform logical reasoning tasks with impressive accuracy \citep{DBLP:journals/corr/abs-2201-11903,DBLP:journals/corr/abs-2204-02311,DBLP:journals/corr/abs-2206-14858}. %suggesting that they are capable of at least some deductive reasoning . %\hh{Not sure if we want to motivate from the spurious correlation angle as we don't have evidence for shortcuts in existing data. Also, we are not in the supervised learning setting so spurious correlation is less of an issue. For CoT, it's more about knowing the conclusion from pretraining vs deducing it from the given facts.}
In CoT prompting, each example consists of a \emph{question} (e.g., ``$\smash{\frac{6}{3}} - 1$?''),
a short description of the reasoning required to answer the question called the \emph{``chain-of-thought''} (e.g., ``$\smash{\frac{6}{3}}$ is $2$. $2 - 1$ is $1$.''), and a \emph{label} (e.g., ``$1$'').
When prompted with a few CoT examples, the elicited reasoning allows LLMs to predict the label with much higher accuracy than standard question-answer prompting.
%The language model is prompted with a sequence of few-shot examples of this format followed by a question. The task is then to predict the CoT and answer label for the test question.
%\hh{The critique of the current results/understanding needs to be more specific. Given that our main contribution is to analyze the model's reasoning chain, here we should also focus on how current eval fails on this.}
However, %prompting / in-context learning is known to be sensitive to a number of variables \hh{cite},
it is unclear to what extent these models can reason due to several confounding factors.
%\hh{This paragraph has too many points. We need to choose one or two most important ones to highlight throughout the paper. The previous paragraph talks about the challenges: 1) using real-world data confounds memorization with reasoning; 2) no easy way to analyze model's CoT. And in this paragraph we should talk about how our solution (the synthetic dataset) addresses the two problems.}
First, 
%There are a number of attempts to measure the reasoning ability of LLMs,
existing studies primarily rely on question-answering (QA) tasks from real-world settings such as math word problems \citep{DBLP:journals/corr/abs-2110-14168,DBLP:journals/corr/abs-2209-00840,DBLP:journals/corr/WestonBCM15}.
It is likely that LLMs have already acquired the knowledge through pretraining and simply retrieve the answer rather than reason over it.
Second, the reasoning task may contain spurious correlations that allow the model to obtain the correct answer through shortcuts \citep{DBLP:journals/corr/abs-2205-11502}.
%To paint a clearer picture of the logical reasoning capability of LLMs,
In this work, we systematically investigate the reasoning capability of LLMs %through controlled studies over
%the boundary between the kinds of reasoning that LLMs are capable of performing and those that LLMs are not.
%One promising approach to evaluate the reasoning capability of LLMs is to
by directly evaluating their predicted chains-of-thought (the interpretable proof steps), rather than the predicted label. %However, chain-of-thought analysis is a difficult task.

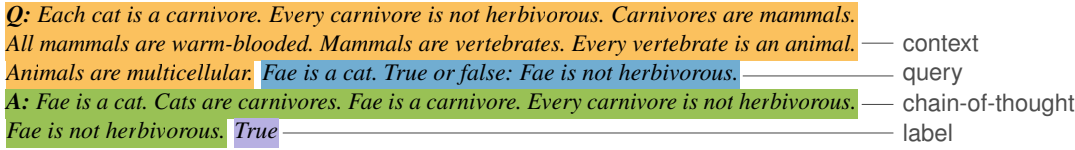
\begin{figure}
    \vspace{-1.5em}
    \centering
    \begin{tikzpicture}
        \definecolor{color2}{RGB}{52, 138, 189}
        \definecolor{color3}{RGB}{152, 142, 213}
        \definecolor{color4}{RGB}{251, 193, 94}
        \definecolor{color5}{RGB}{142, 186, 66}
        \node[text width=32em] (question) {
            \footnotesize\textit{\hlc[color4]{\textbf{Q:} Each cat is a carnivore. Every carnivore is not herbivorous. Carnivores are mammals. All mammals are warm-blooded. Mammals are vertebrates. Every vertebrate is an animal. Animals are multicellular.} \hlc[color2!70]{Fae is a cat. True or false: Fae is not herbivorous.}} \newline
            \textit{\hlc[color5!90]{\textbf{A:} Fae is a cat. Cats are carnivores. Fae is a carnivore. Every carnivore is not herbivorous. Fae is not herbivorous.} \hlc[color3!70]{True}}
        };
        \node[right=1.2em of question,yshift=1.2em] (context) { \color{black!70}\footnotesize\textsf{context} };
        \node[right=1.2em of question,yshift=-0.04em] (query) { \color{black!70}\footnotesize\textsf{query} };
        \node[right=1.2em of question,yshift=-1.12em] (cot) { \color{black!70}\footnotesize\textsf{chain-of-thought} };
        \node[right=1.2em of question,yshift=-2.2em] (label) { \color{black!70}\footnotesize\textsf{label} };
        \draw[draw=black!70] (context.west) -- ([xshift=-1.2em]context.west);
        \draw[draw=black!70] (query.west) -- ([xshift=-5.7em]query.west);
        \draw[draw=black!70] (cot.west) -- ([xshift=-1.2em]cot.west);
        \draw[draw=black!70] (label.west) -- ([xshift=-23.1em]label.west);
    \end{tikzpicture}\vspace{-0.6em}
    \caption{A question-answering example from \dataset{}, with each component highlighted and labeled.}
    \label{fig:example_question}
\end{figure}

To enable easy analysis of the CoT, we construct a new synthetic QA dataset called \dataset{}, for \textbf{Pr}oof and \textbf{Onto}logy-Generated \textbf{Q}uestion-\textbf{A}nswering. Inspired by the \textsc{ProofWriter} dataset \citep{DBLP:conf/acl/TafjordDC21}, each example in \dataset{} is generated from an ontology and has a unique proof (see figure \ref{fig:example_question} for an example).
We convert the proofs into syntactically simple sentences using a grammar such that the inverse process is relatively easy: From the predicted CoT, we semantically parse each sentence into a formal language and reconstruct the underlying proof steps. We then directly analyze the model's reasoning by inspecting each step in the reconstructed proof and comparing them against the gold proof.\footnote{
All analysis code, data, data generation scripts, and model outputs are available at \href{https://github.com/asaparov/prontoqa}{\texttt{github.com/asaparov/prontoqa}}.
} We emphasize here that while the dataset is an important contribution of this paper, the main contribution is the analysis that is facilitated by the dataset.

\begin{figure}
    \centering
    \vspace{-0.6em}
    \tikzset{
    state/.style={
            fill={RoyalBlue!60!black},
            rounded corners=0.2em,
            minimum height=1.2em,
            text width=7em,
            inner sep=2pt,
            text centered,
        },
    multistate/.style={
            rectangle split,
            rectangle split parts=2,
            rectangle split part fill={RoyalBlue!60!black,RoyalBlue!30},
            rounded corners=0.2em,
            minimum height=2.4em,
            text width=7em,
            inner sep=2pt,
            text centered,
        }
    }
    \makebox[\textwidth][c]{\scalebox{0.9}{\begin{tikzpicture}
        \footnotesize
        \node[multistate] (n1) { $\color{white}\texttt{animal}$ \nodepart{two} $\texttt{multicellular}$ };
        \node[state, below=1.3em of n1] (n2) { $\color{white}\texttt{vertebrate}$ };
        \node[multistate, below=1.3em of n2] (n3) { $\color{white}\texttt{mammal}$ \nodepart{two} $\texttt{warm\_blooded}$ };
        \node[multistate, below=1.3em of n3] (n4) { $\color{white}\texttt{carnivore}$ \nodepart{two} $\neg\texttt{herbivorous}$ };
        \node[state, below=1.3em of n4] (n5) { $\color{white}\texttt{cat}$ };
        \draw[-stealth,draw=RoyalBlue!70!black] (n2) -- (n1);
        \draw[-stealth,draw=RoyalBlue!70!black] (n3) -- (n2);
        \draw[-stealth,draw=RoyalBlue!70!black] (n4) -- (n3);
        \draw[-stealth,draw=RoyalBlue!70!black] (n5) -- (n4);
        \node[above=1em of n1, text width=9em] (step1) { \color{black!70}\textsf{Step 1:} \\ \textsf{Generate ontology} };
        
        \node[right=2.6em of step1] (step2) { \color{black!70}\textsf{Step 2: Generate proof from ontology} };
        \node[right=4em of n1, text width=44em] (proof) {
            \footnotesize\prftree[r]{Hop}{
                \prftree[r]{Hop}{
                    \prftree[r]{Ax}{\color{RoyalBlue} \texttt{cat}(\texttt{fae})}
                }{
                    \prftree[r]{Ax}{\color{RoyalBlue} \forall x(\texttt{cat}(x) \to \texttt{carnivore}(x))}
                }{\color{RoyalBlue} \texttt{carnivore}(\texttt{fae})}
            }{
                \prftree[r]{Ax}{\color{RoyalBlue} \forall x(\texttt{carnivore}(x) \to \neg\texttt{herbivorous}(x))}
            }{\color{RoyalBlue} \neg\texttt{herbivorous}(\texttt{fae})}
        };
        
        \node[right=3.8em of n3, yshift=1.8em] (step3) { \color{black!70}\textsf{Step 3: Translate ontology to natural language context} };
        \node[right=5em of n4, text width=32em, yshift=2.8em] (question) { \footnotesize\textit{``\textbf{Q:} Each cat is a carnivore. Every carnivore is not herbivorous. Carnivores are mammals. All mammals are warm-blooded. Mammals are vertebrates. Every vertebrate is an animal. Animals are multicellular.''} };
        
        \node[right=3.8em of n5, yshift=1.8em] (step4) { \color{black!70}\textsf{Step 4: Translate proof into query, chain-of-thought, and label} };
        \node[right=5em of n5, text width=32em, yshift=-0.7em] (cot) {
            \footnotesize\textit{``Fae is a cat. True or false: Fae is not herbivorous.} \newline
            \textit{\textbf{A:} Fae is a cat. Cats are carnivores. Fae is a carnivore. Every carnivore is not herbivorous. Fae is not herbivorous. True''}
        };

        \draw[draw=black!70,-stealth,line width=1.34pt] ([xshift=0.8em]n3.east) -- +(1.2em,0) |- ([xshift=-0.4em,yshift=0.5em]proof.west);
        \draw[draw=black!70,-stealth,line width=1.34pt] ([xshift=0.8em]n3.east) -- ([xshift=-1.4em,yshift=0.79em]question.west);
        \draw[draw=black!70,-stealth,line width=1.34pt] ([xshift=-7em]proof.south east) |- ([xshift=0.7em]cot.east);
    \end{tikzpicture}}}\vspace{-0.4em}
    \caption{Schematic of the generative process for each example in \dataset. \textbf{Step 1:} We generate an ontology from a prior distribution, shown here as a tree. Each node denotes a concept (e.g., \texttt{\color{RoyalBlue}mammal}), each with an optional property (e.g., \texttt{\color{RoyalBlue}warm\_blooded}), and each blue edge denotes a ``subtype of'' relation. \mbox{\textbf{Step 2:}} Generate proof from the ontology. Each horizontal black line indicates a proof step, with its premises written above the line and the conclusion written below. \textbf{Step 3:} Convert the ontology into a natural language context. \textbf{Step 4:} Convert the proof into a natural language query, chain-of-thought, and answer label. There is a one-to-one correspondence between the conclusion of each proof step and the sentences in the chain-of-thought.%\hh{Nice figure!}
    }
    %\hh{We only need one to illustrate the point. But I'd like to add traversing arrows corresponding to an example, perhaps also with the corresponding NL statements.}
    \label{fig:example_generation}
\end{figure}
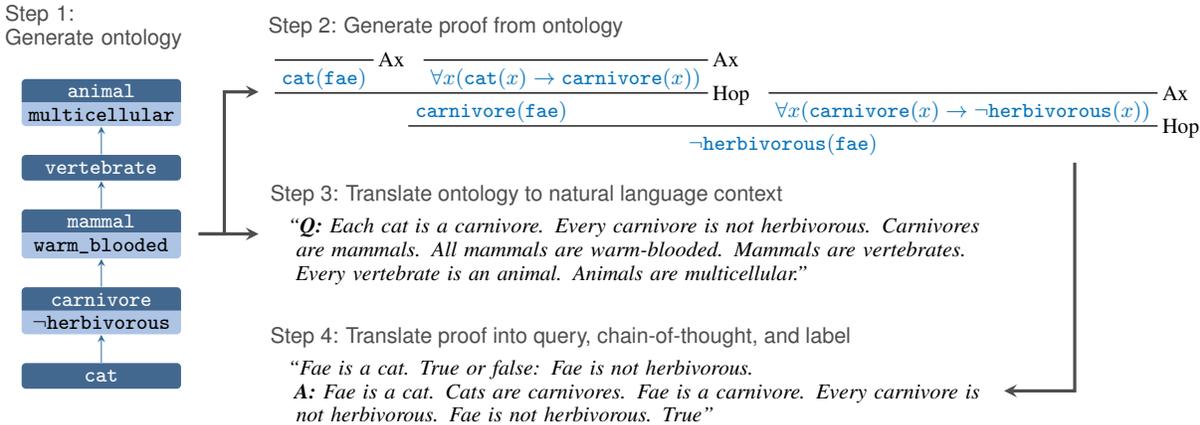

%\hh{We didn't do fine-tuning so let's not mention it in the intro (can mention it in conclusion/discussion).}
%Our generated dataset also can provide new training data to pretrain models to further investigate or possibly improve their ability to reason. %\href{https://github.com/asaparov/synthetic_reasoning_qa}{\texttt{github.com/asaparov/synthetic\_reasoning\_qa}}.
%\begin{itemize}
%    \item Which rules of deduction do LLMs know how to correctly perform? Our initial exploration is guided by the rules of deduction as described by \emph{natural deduction}, a well-studied proof calculus initially developed for mathematics. However, this raises another interesting question: Perhaps a different proof calculus is better suited for LLMs? Perhaps they are more likely to be able to reason using more ``natural'' proof calculi.
%    \item How sensitive is this ability to variables that theoretically should not affect reasoning? For example, variables such as the ordering of sentences, the phrasing of the sentences/question, domain, etc.
%\end{itemize}

%\hh{I find this paragraph not very informative; we don't need to describe the structure of the paper for a 9-page standard submission.}
%\hh{This paragraph should summarize our main results instead, which answers the two questions we started with.}
We systematically evaluate \textsc{InstructGPT}\footnote{\textsc{InstructGPT} is the model resulting from fine-tuning \textsc{GPT-3} via reinforcement learning from human feedback. Throughout the paper, ``\textsc{InstructGPT}'' refers to the model named \texttt{text-davinci-002}. But note that in our experiments, we also evaluate \texttt{text-ada-001}, \texttt{text-babbage-001}, \texttt{text-curie-001}, \texttt{davinci}, and \texttt{text-davinci-001}.} \citep{DBLP:journals/corr/abs-2203-02155} and the original \textsc{GPT-3} \citep{DBLP:conf/nips/BrownMRSKDNSSAA20} on \dataset{} by controlling a number of variables that characterize the complexity of the reasoning task, such as the ontology type and the number of proof steps required.
Our analysis shows that these models are quite good at producing {valid} \emph{individual} proof steps, even on fictional and counterfactual ontologies.
However, LLMs have difficulty with \emph{proof planning}: when the models encounter a point in the proof where
multiple valid proof steps are available, they sometimes select the wrong step, and this often leads to an incomplete proof and subsequently an incorrect answer.
Interestingly, the models are much less likely to be misled with a true ontology, %which leads to much higher accuracies
suggesting that the world knowledge acquired during pretraining plays an important role in LLM reasoning.
We also find that our results generalize to more sophisticated/informative prompts, such as \emph{self-consistency} prompting \citep{DBLP:journals/corr/abs-2203-11171}, and prompts with example traces of depth-first proof search instead of CoT.
%We also find that proof accuracy is quite well-correlated with label accuracy, indicating that label accuracy is a good measure for the reasoning ability of LLMs, provided that sufficient care is taken to mitigate the effect of heuristics and spurious correlations.
%This work is structured into two main parts:
%\begin{itemize}
%    \item In the first part, we introduce the data and the task in section \ref{sec:synthetic_qa_task}, describe our method to evaluate the correctness of proofs in section \ref{sec:evaluating_proofs}, which we utilize to systematically evaluate the reasoning ability of \textsc{InstructGPT} and \textsc{GPT-3} in section \ref{sec:proof_analysis}.
%    \item In the second part, in section \ref{sec:error_analysis}, we further analyze the errors produced by \textsc{InstructGPT} and \textsc{GPT-3} on the reasoning task and identify the primary cause of the errors in reasoning. We then suggest ways in which these limitations can be addressed to possibly improve the reasoning ability of LLMs.
%\end{itemize}

\section{Related work}

%\hh{The related work should not be described at the level of individual papers. It is supposed to situate the work in a much broader context by giving an overview of each related direction. Think of it as a mini-survey. In addition, for each direction, we need to describe how our work is related to or differentiated from it. Example: see section 7 here: https://arxiv.org/pdf/2208.01066.pdf.}

%\hh{Specifically, for this work, I think there are several related directions:
%1) CoT prompting.
%2) Benchmarks for logical reasoning: datasets.
%3) Reasoning in natural language: Greg and Yejin's work.
%4) Symbolic reasoning: semantic parsing, neurosymbolic systems.
%}

%\hh{If you have time, you can try to describe the *contribution* (i.e. what they find, what's the significance of this piece of work in a larger context, how do everything connects) of these work rather than just what they did.
%If someone wants to learn about the state of the field, I don't think they get much away from the current description.
%}

Our proposed dataset is most closely related to \textsc{ProofWriter} \citep{DBLP:conf/acl/TafjordDC21} and \textsc{FOLIO} \citep{DBLP:journals/corr/abs-2209-00840} which are QA datasets designed to test reasoning ability.
\textsc{ProofWriter} provides multi-hop proofs for each example.
However, there are a number of key properties that led us to develop our own dataset (see table \ref{tab:datasets} for a summary).
\textsc{FOLIO} does not provide easily-parseable proofs/CoTs in their examples, and evaluation is done by inspecting the predicted labels, which may not necessarily be a good measure of reasoning ability.
In our analysis, we focus on more specific variables that may affect the reasoning of the model, such as: (1) Is the model's reasoning dependent on whether the example is consistent with pretraining (``true''), inconsistent (``false''), or neither (``fictional'')? (2) Is the model's reasoning sensitive to whether the predicates in the examples are unary or binary? (3) Is the model's reasoning dependent on the rules of deduction in the examples?
These variables are not controllable in existing datasets.
Further, in some datasets, the code to generate examples is not available. %and our exploration would benefit from the flexibility and controllability that is afforded by working with our own dataset, such as the ability to add distractors to mitigate heuristics.

There are efforts to tweak or extend CoT prompting to elicit more sophisticated reasoning behavior \citep{DBLP:journals/corr/abs-2205-09712,DBLP:journals/corr/abs-2203-11171,DBLP:journals/corr/abs-2208-14271,DBLP:journals/corr/abs-2207-04901,DBLP:journals/corr/abs-2207-10342}, and they have shown that these prompting extensions to CoT can improve the elicited reasoning behavior of LLMs, even with smaller models.
Rather than presenting a new prompting approach, the goal of this work is to measure the reasoning ability elicited by CoT.
There are other datasets that have been designed or used to measure the reasoning capabilities of transformer-based models and LLMs \citep{DBLP:journals/corr/abs-2209-00840,DBLP:journals/corr/WestonBCM15}.
They show that LLMs are able to answer questions that require reasoning in the few-shot setting with reasonable accuracy.
Similar to our approach, \citet{DBLP:journals/corr/abs-2009-07185} converts logical forms into fairly simple natural language using templates.
However, the examples in these datasets are consistent with the real-world, and so they may confound measuring reasoning ability with retrieval ability.
%In addition, they evaluate the predicted labels, which may not necessarily be a good measure of reasoning ability.
\citet{DBLP:journals/corr/abs-2206-10498} found that LLMs had difficulty with a fairly simple planning task, but it is not clear whether this was due to an inability to reason or other abilities instrumental in planning, such as world modeling, keeping track of state changes, and reasoning about events that occur sequentially in time. This is despite their controlling for other variables involved in planning, such as plan generation, robustness to goal formulation, among others. They experimented with examples in a ``Blocksworld'' environment, a significant portion of which the model can acquire from pretraining. Our work aims to address this gap.
As in our approach, \citet{DBLP:journals/corr/abs-2207-07051} specifically looked at whether LLMs can reason in fictional or counterfactual settings and found that reasoning ability is indeed negatively affected in these settings. However they did not analyze individual steps of reasoning to better understand the cause of the errors.
Since we are able to formally evaluate the LLM's predicted CoT step-by-step, we are able to perform a more fine-grained analysis of their reasoning ability.
\citet{DBLP:journals/corr/abs-2205-11502} showed that \textsc{BERT} is not able to learn to reason robustly, but they did not use CoT prompting and it is not obvious if their results generalize to LLMs, which we evaluate.

\begin{table}
    \vspace{-1em}
	\scriptsize
	\sffamily
	\def\arraystretch{1.15}
	\setlength{\tabcolsep}{2pt}
	\hspace{-1.4em}\begin{tabular*}{\textwidth}{|m{0.156\textwidth}|>{\centering\arraybackslash}m{0.13\textwidth}|>{\centering\arraybackslash}m{0.13\textwidth}|>{\centering\arraybackslash}m{0.13\textwidth}|>{\centering\arraybackslash}m{0.13\textwidth}|>{\centering\arraybackslash}m{0.14\textwidth}|>{\centering\arraybackslash}m{0.13\textwidth}|}
		\hhline{|-|------|}
		\centering\arraybackslash\textbf{Dataset} & Provides easily semantically-parseable proofs & Controls for true vs false vs fictional contexts & Controls for unary vs binary predicates & Controls for specific rules of deduction &Tests reasoning beyond the domain of math word problems & Generation code available \\
		\hhline{:=:======:}
		\texttt{GSM8K} \linebreak \citet{DBLP:journals/corr/abs-2110-14168} &
			\cellcolor{pale_red!35} \xmark &
			\cellcolor{pale_red!35} \xmark &
			\cellcolor{pale_red!35} \xmark &
			\cellcolor{pale_red!35} \xmark &
			\cellcolor{pale_red!35} \xmark &
			\cellcolor{pale_yellow!35} human-annotated \\
		\hhline{|-|------|}
		\texttt{ProofWriter} \linebreak \citet{DBLP:conf/acl/TafjordDC21} &
			\cellcolor{pale_green!35} \cmark &
			\cellcolor{pale_red!35} \xmark &
			\cellcolor{pale_red!35} \xmark &
			\cellcolor{pale_yellow!35} $\bm{\sim}$ &
			\cellcolor{pale_green!35} \cmark &
			\cellcolor{pale_red!35} \xmark \\
		\hhline{|-|------|}
		\texttt{FOLIO} \linebreak \citet{DBLP:journals/corr/abs-2209-00840} &
			\cellcolor{pale_red!35} \xmark &
			\cellcolor{pale_red!35} \xmark &
			\cellcolor{pale_red!35} \xmark &
			\cellcolor{pale_red!35} \xmark &
			\cellcolor{pale_green!35} \cmark &
			\cellcolor{pale_yellow!35} human-annotated \\
		\hhline{|-|------|}
		\texttt{SimpleLogic} \linebreak \citet{DBLP:journals/corr/abs-2205-11502} &
			\cellcolor{pale_red!35} \xmark &
			\cellcolor{pale_red!35} \xmark &
			\cellcolor{pale_green!35} \cmark &
			\cellcolor{pale_green!35} \cmark &
			\cellcolor{pale_green!35} \cmark &
			\cellcolor{pale_green!35} \cmark \\
		\hhline{:=:======:}
		\dataset{} \linebreak (proposed dataset) &
			\cellcolor{pale_green!35} \cmark &
			\cellcolor{pale_green!35} \cmark &
			\cellcolor{pale_green!35} \cmark &
			\cellcolor{pale_green!35} \cmark &
			\cellcolor{pale_green!35} \cmark &
			\cellcolor{pale_green!35} \cmark \\
		\hhline{|-|------|}
	\end{tabular*}
	\caption{Comparison of existing datasets for the formal analysis of reasoning ability.}\label{tab:datasets}
	\vspace{-2ex}
\end{table}

There are two broad research approaches for reasoning in NLP: (1) reasoning over a formal symbolic language, possibly with neuro-symbolic methods and/or semantic parsing \citep{DBLP:journals/tacl/SaparovM22,Zhang2022,DBLP:conf/acl/KapanipathiARRG21,DBLP:conf/iclr/DongMLWLZ19,DBLP:conf/nips/Rocktaschel017}, or (2) reasoning directly over natural language \citep{DBLP:conf/emnlp/ChenCD21,DBLP:journals/corr/abs-2201-06028,DBLP:conf/emnlp/BostromZCD21,DBLP:conf/nips/Welleck0BHCCC21,DBLP:conf/iclr/BhagavatulaBMSH20,DBLP:conf/emnlp/AngeliM14,DBLP:conf/iwcs/MacCartneyM09}. While \dataset{} is generated from symbolic ontologies, the examples themselves are in natural language, and so provides value to both research directions. %\abucomment{not sure what sources to cite here; some of the earlier papers fall into this direction, but we could make a huge list for both} \abucomment{also, am i missing any important papers from Greg/Yejin?}

Recent work has examined in-context learning and found that performance on certain tasks is sensitive to the prompt \citep{DBLP:journals/corr/abs-2202-07206,DBLP:conf/acl/LuBM0S22}. However, they focused on sentiment classification and simple arithmetic tasks, and it is not clear if their results generalize to reasoning. %It is very possible that LLMs have learned to better generalize using rules of deduction such as modus ponens. Furthermore,
The LLM could feasibly use retrieval, rather than reasoning, to perform those tasks. Our experiments on the fictional ontology show that the model is able to reason even when there is nothing to retrieve from.

\section{\dataset: A synthetic dataset for logical reasoning} \label{sec:synthetic_dataset}
%\vspace{-1.4em}

%\hh{The description of the dataset is procedural now and we need to make it more structured and modular. I tried to add some paragraph titles below. Can we restructure the content along these lines?}

%\hh{First motivate why we focus on modus ponens.}

%Formal analysis of the reasoning capability of LLMs is greatly facilitated by the ability to parse the sentences in the predicted CoT into unique symbolic logical forms and interpret them as steps in a larger proof.
We create a new dataset, called \dataset{} for \textbf{Pr}oof and \textbf{Onto}logy-Generated \textbf{Q}uestion-\textbf{A}nswering, where each question is generated from a symbolic ontology and proof
to facilitate formal analysis of the predicted CoT.
To focus the scope of our exploration, and to limit the complexity of the generated questions to those within reach of current LLMs, we only consider questions that are answerable using repeated applications of the modus ponens deduction rule.
More formally, \emph{modus ponens} is a simple deduction rule where given the premises $\color{RoyalBlue}\forall x(f(x) \to g(x))$ and $\color{RoyalBlue}f(a)$ , we conclude $\color{RoyalBlue}g(a)$ (e.g., given ``All cats are carnivores'' and ``Fae is a cat,'' we conclude ``Fae is a carnivore;'' see figure \ref{fig:restricted_proof_calculus} in the appendix).\footnote{In natural deduction, this rule is actually a composition of two steps: given $\forall x(f(x) \to g(x))$, use universal elimination to conclude $f(a) \to g(a)$, and given $f(a)$, use implication elimination to conclude $g(a)$.}
This rule can be easily chained together to construct proofs with controllable size.

We generate CoT examples
consisting of: the context, query, CoT, and label,
where the \emph{context} is a short paragraph containing information relevant to answer the \emph{query} (see figure \ref{fig:example_question} for an example).
%\hh{Can we label the parts in figure 1 or show an example here with the parts labeled.}
Each example is translated from a proof and ontology
such that the inverse process is simple: the sentences in an example can be easily and uniquely parsed into symbolic logical forms amenable to formal analysis.
%Each step in the proof is translated into a sentence in the CoT using a simple grammar, so that the inverse process is simple: the output sentences can be easily and uniquely parsed into symbolic logical forms amenable to formal analysis.
%But if the generated questions are too difficult for the LLMs to answer, we would not be able to shed light on their reasoning ability.
%\hh{I edited it a bit. Please check that everything is right.}
More specifically, as shown in figure \ref{fig:example_generation}, we: (1) first generate an ontology from a set of concepts, (2) generate a proof by traversing the ontology, (3) translate the ontology into the natural language context, and (4) translate the proof into the query, CoT, and label by mapping logical forms to natural language sentences.
We describe each step in further detail below.

\paragraph{Ontology generation.}
The first step is to generate a small hierarchical \emph{ontology}. The ontology is a set of concepts (e.g., \texttt{\color{RoyalBlue}mammal}, \texttt{\color{RoyalBlue}cat}, \texttt{\color{RoyalBlue}carnivore}, etc) and subtype relations between them (e.g., $\color{RoyalBlue}\forall x(\texttt{cat}(x) \to \texttt{carnivore}(x))$). The ontology also describes properties of concepts (e.g., $\color{RoyalBlue}\forall x(\texttt{mammal}(x) \to \neg\texttt{cold\_blooded}(x))$).
To generate questions that are not overly complex, we restrict the ontologies to be \emph{linear} (i.e., in the tree, every node has exactly 0 or 1 child nodes).
Since ontologies are randomly generated, they vary in size from as few as $3$ concepts to as many as $10$.

\paragraph{Proof generation.}
We generate proofs from the ontology by choosing a starting node uniformly at random, and generating the initial axiom indicating that an entity has a specific type (e.g., $\color{RoyalBlue}\texttt{cat}(\texttt{fae})$). Then, we walk up the tree, with each step corresponding to an application of a deduction rule (i.e., a \emph{proof step}).
Each proof step consists of zero or more \emph{premises} and one \emph{conclusion}.
We stop when we reach a node (e.g., $\color{RoyalBlue}\texttt{carnivore}(\texttt{fae})$), or a node property (e.g., $\color{RoyalBlue}\neg\texttt{herbivorous}(\texttt{fae})$), such that the number of generated proof steps matches the target number of steps.
%To control the complexity of the generated questions, we restrict the reasoning to two rules of deduction, detailed in figure \ref{fig:restricted_proof_calculus}.\hh{This sentence should be after "... an application of a deduction rule". Actually it may be redundant, as we have described the two deduction rules in the first para of this section.}

\paragraph{Translation to natural language example.}
%\hh{Should we call it CoT generation since it's more than the question?}
Given a generated ontology and proof, we now translate it into a natural language CoT example consisting of the question (context and query), CoT, and label. We describe how each component is generated below:

We use a simple grammar to convert the formal statements of the ontology into the natural language utterances that make up the context. Every edge in the ontology is converted into sentences such as ``All cats are carnivores'' or ``Every cat is a carnivore.'' Properties of nodes are also converted into sentences of the form ``All mammals are not cold-blooded,'' etc.

%\hh{This part (question generation) needs to be made more clear. At this point, as a reader, I know that a bunch of NL statements is generated from the ontology. But where does the proof come from? We need to say upfront that given a generated proof, we are now going to translate it into a CoT example consisting of question (context + query), CoT, answer. And then how each part is generated.}
The query is generated by using the same grammar to convert the initial axiom in the proof into a natural language sentence (e.g., ``Fae is a cat''). We then determine with probability 0.5 whether to ask if the conclusion of the proof is true or if its negation is false, and convert it into a natural language ``true or false'' query (e.g., ``True or false: Fae is not herbivorous.'') and label (e.g., ``True'').

We convert the ordered sequence of proof steps into the CoT by translating the conclusion of each proof step into a CoT sentence.

\paragraph{Avoiding shortcuts.} In section \ref{sec:avoiding_shortcuts} in the appendix, we describe how we add \emph{distractor sentences} in order to remove shortcuts that would allow the model to ``guess'' the answer without reasoning.

A unique feature of \dataset{} is that it is easily programmable, with a handful of tunable knobs which we use to generate examples with varying degrees of complexity and study different aspects of reasoning in LLMs. These variables are described in greater detail in section \ref{sec:experimental_setup}.

%\begin{figure}
%    \raggedright
%    \hspace{0.05\textwidth}\begin{minipage}{0.9\textwidth}
%        \textbf{Question text:} \textit{``Each cat is a carnivore. Every carnivore is not herbivorous. Carnivores are mammals. All mammals are warm-blooded... Fae is a cat. True or false: Fae is not herbivorous.''} \\[0.5em]
%        \textbf{Chain-of-thought:} \textit{``Fae is a cat. Cats are carnivores. Fae is a carnivore. Every carnivore is not herbivorous. Fae is not herbivorous. True''} \\[0.5em]
%        \textbf{Reconstructed proof:}
%    \end{minipage} \\[-0.7em]
%    \begin{equation*}
%        \footnotesize\prftree[r]{Hop}{
%            \prftree[r]{Hop}{
%                \prftree[r]{Ax}{\color{RoyalBlue} \texttt{cat}(\texttt{fae})}
%            }{
%                \prftree[r]{Ax}{\color{RoyalBlue} \forall x(\texttt{cat}(x) \to \texttt{carnivore}(x))}
%            }{\color{RoyalBlue} \texttt{carnivore}(\texttt{fae})}
%        }{
%            \prftree[r]{Ax}{\color{RoyalBlue} \forall x(\texttt{carnivore}(x) \to \neg\texttt{herbivorous}(x))}
%        }{\color{RoyalBlue} \neg\texttt{herbivorous}(\texttt{fae})}
%    \end{equation*}
%    \vspace{-1.5em}
%    \caption{An example of a chain-of-thought for a given question, and its reconstructed proof. In this example, the proof contains 2 hops.}
%    \label{fig:example_proof}
%\end{figure}

\section{Formal analysis of predicted proofs} \label{sec:evaluating_proofs}
\vspace{-0.1em}
Instead of measuring the accuracy of the predicted answers (i.e.,\ ``true'' or ``false''),
we would like to directly evaluate the predicted CoT to check if the model derives the right answer for the right reason.
%\hh{Important: it is unclear from the description whether the evaluation operates over each step in the reconstructed / gold proofs or each sentence in the predicted / gold CoTs.}
%\hh{Can we highlight the point that we are evaluating the correctness of every proof step?}
We endeavor to analyze whether the model is able to apply deduction rules correctly at each proof step (i.e., local correctness), but also whether the model can plan ahead and work toward proving the answer for the query (i.e., global correctness).
To measure the local correctness of a given proof step, we compute whether the step follows from one or more applications of deduction rules, and whether it requires additional rules beyond those of the gold proofs. %listed in figure \ref{fig:restricted_proof_calculus}\hh{If we decide to move the figure to appendix, then let's not refer to it frequently (which makes it essential in the main text).}.
To measure the global correctness, we wish to identify proof steps that deviate from the gold proof.
%\hh{We are jumping into the details too fast. We have just said that we want to evaluate the CoT to check the correctness of the reasoning. How are we going to check it? We will check whether each sentence can be proved from the previous ones and whether it deviates from the gold proof (i.e. proving irrelevant conclusions). (Actually the local/global correctness in the next para are good high level motivation here). Note that up this point we don't need to invoke the concept of proof step and we are just describing the intuition (i.e. how a human would check it). Then, only in the next para when we talk about how we implement this idea do we need to talk about logical forms and proof steps.}

\hspace{-0.4em}\begin{table}
    \footnotesize
    \vspace{-1.8em}
 	\begin{tabular}{ >{\raggedleft\arraybackslash}m{0.24\textwidth} | m{0.61\textwidth} } \toprule[0.1em]
		\textbf{Step type} & \textbf{Example} (the conclusion of each step is highlighted green) \\ \midrule[0.1em]
		Strictly-valid atomic correct step, \newline or \emph{canonical step} & \textit{``Fae is a cat. Cats are carnivores. Fae is a carnivore. \textcolor{OliveGreen}{Every carnivore is not herbivorous.} Fae is not herbivorous. True''} \newline \hspace*{\fill} \textcolor{RedOrange!90!black}{\textbf{(this is the gold CoT for this example)}} \\ \midrule
		Strictly-valid atomic misleading step & \textit{``Fae is a cat. Cats are carnivores. Fae is a carnivore. \textcolor{OliveGreen}{Every carnivore is a mammal.} Fae is a mammal...''} \\ \midrule
		Strictly-valid non-atomic correct step & \textit{``Fae is a cat. \textcolor{OliveGreen}{Fae is a carnivore.} Every carnivore is not herbivorous. Fae is not herbivorous. True''} \\ \midrule
		Strictly-valid non-atomic misleading step & \textit{``Fae is a cat. Cats are carnivores. Fae is a carnivore. \textcolor{OliveGreen}{Fae is a mammal.} Every mammal is a vertebrate...''} \\ \midrule
		Broadly-valid correct step & \textit{``Fae is a cat. \textcolor{OliveGreen}{Every cat is not herbivorous.} Fae is not herbivorous...''} \\ \midrule
		Broadly-valid misleading step & \textit{``Fae is a cat. \textcolor{OliveGreen}{Every cat is a mammal.} Fae is a mammal...''} \\ \midrule
		Invalid step & \textit{``Fae is a cat. Cats are carnivores. Fae is a carnivore. \textcolor{OliveGreen}{Every carnivore is a cat.} Fae is a cat...''} \\ \bottomrule[0.1em]
	\end{tabular}\vspace{-0.4em}
    \caption{The types of proof steps (and examples thereof) into which we categorize each step in the predicted chain-of-thought from LLMs. Compare the given chain-of-thought examples with the gold example provided in the first row.
    %\hh{Can we copy the gold CoT to the table? Actually, it's just the first row, right? Can we indicate it somehow}
    %\hh{Can we also underline the part of text that corresponds to the proof step?}
    }
    \label{fig:proof_step_types}
\end{table}

%\hh{It's a strong statement to say that LLMs reason like humans etc. We can state this as a post-hoc conjecture, but not as a motivation of our analysis.}
%\hh{We should also describe the high level motivation for the analysis (i.e. what are we trying to measure; how would one come up with these dimensions from scratch?). Maybe something like: locally, we would like to know whether each step of the generated proof is valid - this suggests whether the model is using some deduction rules systematically. Then, we introduce validity and atomicity. Next, globally, we want to know if each step is relevant (is there a better word?) and builds towards the proof target - this measures how well the model plans the proof. Then, we introduce misleading.}
%Some proof steps may still be locally-correct or valid, but require more than one application of the deduction rules, or additional rules beyond those listed in figure \ref{fig:restricted_proof_calculus}.
%However, language models, just like humans when they write down their reasoning, are prone to skip steps or use deduction rules beyond those listed in figure \ref{fig:restricted_proof_calculus}.
%Humans regularly omit steps that can be easily inferred from context for the sake of parsimony (i.e., to conserve energy). Therefore, we categorize each predicted proof step according to three dimensions:
\vspace{-1.4em}
To achieve this, we parse each sentence of the predicted CoT into logical form via recursive-descent parsing using the simple grammar from the generative process. We then compute whether that logical form is provable from previous logical forms via one or more applications of deduction rules. This logical form corresponds to the conclusion of a proof step. We then evaluate the correctness of each proof step by categorizing it according to three dimensions:
%We consider a predicted proof to be correct if it contains a subset of proof steps that constitute the full gold proof. %\hh{The description is mixing sentences and proof steps. It sounds like we just need to check if the sentences are substrings of the gold CoT.}.
%To this end, we categorize each predicted proof step according to three dimensions:
\begin{enumerate}[noitemsep,topsep=-0.3pt,leftmargin=*]
    \item \textit{Validity:} Is the current proof step provable from previous steps? If it is provable using only the deduction rules that appear in the gold proofs, we say the step is \emph{strictly-valid}. If it is provable with a more powerful proof calculus, like natural deduction, we say the step is \emph{broadly-valid}. Otherwise, we say the step is \emph{invalid}.

    For example, given the premises, ``Cats are carnivores'' and ``Carnivores are mammals,'' the step with conclusion ``Cats are mammals'' is broadly-valid since an additional deduction rule is required to prove it: given $\color{RoyalBlue}\forall x(f(x) \to g(x))$ and $\color{RoyalBlue}\forall x(g(x) \to h(x))$, conclude $\color{RoyalBlue}\forall x(f(x) \to h(x))$. Notice that this is distinct from a strictly-valid non-atomic step since this conclusion is not provable via repeated applications of modus ponens.We note that this the only additional rule that we check, as we did not encounter any instances of other broadly-valid rules. %\hh{Is this the only broadly valid rule we consider? If so, we should probably say it.}

    \item \textit{Atomicity:} Is the current proof step provable from previous steps with \emph{exactly one} application of a deduction rule? If so, we say the proof step is \emph{atomic}. Otherwise, it is \emph{non-atomic}. Note that since all broadly-valid steps are non-atomic, this distinction is only useful for strictly-valid steps.

    For example, given the premises, ``Fae is a cat,'' ``Cats are carnivores,'' and ``Carnivores are mammals,'' the step with conclusion ``Fae is a mammal'' is non-atomic since the step ``Fae is a carnivore'' was skipped.

    %For example, given the same premises as above, ``Fae is a carnivore'' is atomic, and ``Fae is a mammal'' is non-atomic.\hh{Can we just explain the non-atomic case and explicitly state the missing step?}

    \item \textit{Utility:} If the current proof step's premises are part of the gold proof, but its conclusion is not, then we say the proof step is \emph{misleading}. Otherwise, it is \emph{correct}.
    %\hh{I'm wondering if we can call it "utility" to be more consistent with the names of the previous two criteria.}

    For example, given the premises ``Fae is a carnivore,'' ``All carnivores are not herbivorous,'' and ``Carnivores are mammals,'' and the goal is to prove ``Fae is not herbivorous,'' the step ``Fae is a mammal'' is misleading since although the step is strictly-valid, it does not help to prove the goal.
    %\hh{Tbh, these examples don't make it easier than the table. It may be trivial to you but we have to think in the reader's shoes. We don't necessarily need to repeat the example, but only need to explain why it's misleading, explicitly. e.g., in this example, "fae is a mammal" is provable but it doesn't help us reach the goal faster.}
    %\abucomment{perhaps i should just refer to the table for the premises? i thought having them more immediately available to the reader might help}
\end{enumerate}
The types of proof steps are listed in table \ref{fig:proof_step_types} along with examples.
Unparseable proof steps are marked as incorrect.
For brevity, we refer to strictly-valid atomic correct steps as \emph{canonical steps}.
Psuedocode of the procedure to evaluate proofs is given in algorithm \ref{alg:evaluate_cot} in the Appendix.

\paragraph{Metrics.}
%\hh{Let's not mix method and results. This section is about the analysis method, so here we should only talk about metrics.}
%\hh{Start by explicitly stating our goal here or the problem we want to address. And we should have a high level plan: we will define "correctness" of each proof step, and compute it's accuracy. The key challenge is what counts as "correct".}
Given the above categorization of proof steps, a proof is defined to be correct if and only if there exists a path of proof steps from the premises to the conclusion (note that under this definition, it is possible for a correct proof to contain invalid proof steps). We could require that all proof steps in the path be canonical. %\hh{Why a subset? If every step is SVAC, then will the proof have non-gold steps (which would make it misleading)?}\abucomment{the proof would still be ``correct'' in this case even if it has misleading steps}
But it is not obvious that this metric, which we call \emph{strict proof accuracy}, would accurately measure the reasoning ability of the model.
%But it is not obvious how to measure the accuracy of a predicted CoT. In a strict setting, we would only count a CoT to be correct if there exists a subset of strictly-valid atomic correct steps that constitute the gold proof. More precisely, the proof is correct if for every step in the gold proof, there exists a corresponding step in the predicted CoT, and the premises precede this step in the CoT. But since LLMs may skip steps in a manner similar to humans, this strict metric may not be the most appropriate to measure the reasoning ability of the model.
As such, we also consider more relaxed metrics: (a) we allow proof steps in the path to be strictly-valid non-atomic correct, which we call \emph{``skip'' proof accuracy}, (b) we allow proof steps to be broadly-valid, which we call \emph{broad proof accuracy}, or (c) we allow proof steps to be strictly- or broadly-valid, which we call \emph{valid proof accuracy}.
%\hh{Not sure what's the point of the last sentence on ordering. How is it related to Metrics?}
%The ordering of the sentences in the CoT is important: the premises of each proof step must precede the conclusion. If not, then we consider that step to be incorrect.
%\hh{Can we name the three metrics? Suggestions: strict accuracy, skip accuracy, broad accuracy, valid accuracy. Depending on space, we might include a table to show what kind of steps each metric allow.}
%\abucomment{TODO: if space allows, add a table outlining these metrics}

\vspace{-0.2em}
\section{Results}

\vspace{-0.2em}
\subsection{Experimental setup} \label{sec:experimental_setup}
\vspace{-0.2em}

%\hh{Need citation for GPT3 and INstructGPT}

%\hh{Need an experiment setup section/paragraph to describe the models, how we run all our experiments, the error bars, decoding algorithm etc., so that we don't repeat them for each experiment.}
In each experiment, we generate QA examples, perform CoT prompting on the LLMs, and analyze the predicted CoTs.
We run the experiments on %\hh{It's unclear what analysis we are running. We need to first say for each experiment, we generate data, perform CoT prompting on GPT, and analyze the generated CoTs. Then, talk about details for each step.}
\textsc{InstructGPT} and the original \textsc{GPT-3} (OpenAI models \texttt{text-ada-001}, \texttt{text-babbage-001}, \texttt{text-curie-001}, \texttt{davinci}, \texttt{text-davinci-001}, \texttt{text-davinci-002}), with greedy decoding \citep{DBLP:journals/corr/abs-2203-02155,DBLP:conf/nips/BrownMRSKDNSSAA20}.
We use 8-shot in-context learning, so each input to the LLM consists of 8 fully-labeled questions followed by a single test question with missing CoT and label.
The model's task is to predict the CoT and label for the test question.
Note that all examples across all inputs are independently and identically generated from \dataset{}.
%\hh{Need to explain the CoT: how many examples in the demonstration? Are they generated in the same way as the test example? Do we use the same demonstration for all test examples?}

%\hh{I wonder if we should move this part to 5.3 because it's more relevant there. The readers may have forgotten about these variables when they get to 5.3. And it's natural to talk about the motivation of the experiment there. }
%\hh{Again, the variables are very well documented, but why are they interesting? We should the reader know about them? We should say something like: A unique feature of our dataset is that the complexity of the reasoning task can be easily controlled. Then we explain why controlling these variables may reveal interesting results and how we control them.}
There are a number of variables that we control when generating examples in \dataset{}: (1) the number of hops, (2) the ordering in which the sentences are generated from the ontology, and (3) the type of the ontology. The number of hops directly controls the difficulty of the generated example, and we experiment with 1, 3, and 5 hops.

We control the ontology traversal direction: We either traverse the tree top-down (i.e., preorder) or bottom-up (i.e., postorder), generating a sentence for each traversed edge/node. The ordering also affects the difficulty of the generated example: if the sentences are generated bottom-up, they will follow the same order as the steps in the gold proof. On the other hand, if they are generated top-down, the order is reversed, and the task may be more difficult.

To avoid any confounding effects from knowledge acquired during pretraining, \dataset{} generates examples with fictional concept names (e.g., ``wumpus'' instead of ``cat,'' etc). But we are also interested in measuring this confounding effect, and so in addition to fictional ontologies, we also generate ``true'' and ``false'' ontologies. True ontologies use real concept names and are consistent with the real-world (we randomly sample from a list of three hand-coded real ontologies). False ontologies use real concept names but the trees are generated using the random process described in section \ref{sec:synthetic_dataset}, and so it is very likely to generate a false statement, such as ``All mammals are cats.''

For each combination of variables, we run the model on 400 examples generated from the testbed, for a total of 48 experiments. We compute 95\% confidence intervals for each experiment, as the number of correct proofs is distributed as $\text{Binomial}(400,p)$ with $p$ being the model's accuracy \citep{Wilson1927-ss}. %\hh{How do we compute the ci? Do we use bootstrap or run multiple random seeds or sth else? Need to explain.}
%\abucomment{is this what you meant by explaining how to compute ci? i can go into further detail here but didn't feel it was necessary, but perhaps readers won't be as familiar?}
%\hh{I see. I was thinking non-parametric methods. Then I guess it's sufficient to just say something like we compute CI of accuracy assuming a binomial distribution.}

\subsection{Do correct answers imply correct reasoning?}

\begin{figure}
    \centering
    \vspace{-1.2em}
    \hspace{-1em}
    \makebox[\textwidth][c]{
        \includegraphics[scale=0.66]{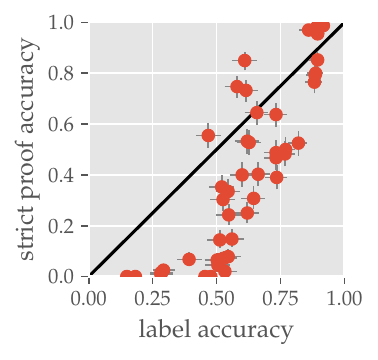}
        \hspace{-0.7em}\includegraphics[scale=0.66]{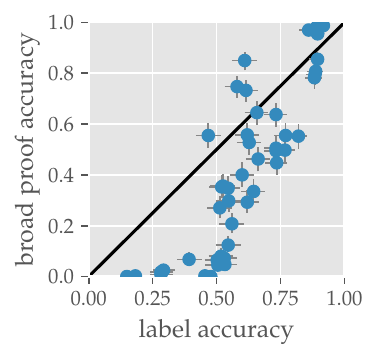}
        \hspace{-0.7em}\includegraphics[scale=0.66]{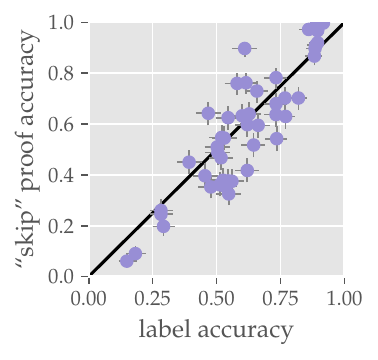}
        \hspace{-0.7em}\includegraphics[scale=0.66]{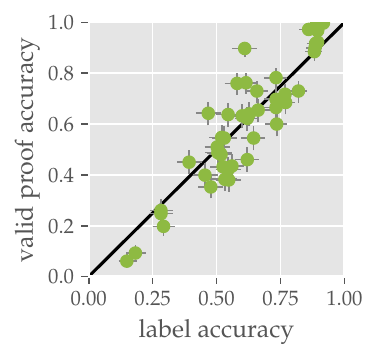}
    }
    \vspace{-0.6em}
    \caption{Scatter plots of label accuracy vs proof accuracy of all \textsc{GPT-3} experiments in this paper. The black line indicates perfect agreement between label accuracy and proof accuracy. We emphasize that ``proof accuracy'' indicates the fraction of proofs (not proof steps) that are considered correct according to our metrics. Label accuracy is not well-correlated with strict or broad proof accuracy, and is better correlated with ``skip'' and valid proof accuracy, suggesting that label accuracy is a good measure of reasoning ability.}
    \label{fig:label_vs_proof_accuracy}
    \vspace{-0.2em}
\end{figure}

%\hh{Why do we want to investigate this question? Need some context here. We have it in the intro but let's remind the readers, plus they may not read the intro and we want to each section to be relatively self-contained.}
Label accuracy may not necessarily measure whether the model is performing reasoning correctly, since the model may find ways to guess the label via heuristics.
To gauge whether label accuracy is a good metric %\hh{Can we add why this might not be a good metric first}
and which proof accuracy metric is best to measure reasoning ability, we investigate how label accuracy is related to the various proof accuracy metrics.
%\abucomment{i could also add the motivation of checking which metric for proof accuracy is best for measuring reasoning ability. should i?}\hh{I think that's fine since it's secondary here?}
%\hh{Also this corresponds to a separate question which is whether the answer accuracy is correlated to proof accuracy.}
We plot proof accuracy vs label accuracy (i.e., simply checking whether the predicted label ``True'' or ``False'' is correct) of every experiment that we conducted in figure \ref{fig:label_vs_proof_accuracy}. Each point in the scatter plot corresponds to one of our 48 experiments described above.
%\hh{Better to make "label accuracy" the subject (rather than proof accuracy) because it is the subject of study here---we want to know if it implies right reasoning. e.g. label accuracy is correlated with...}
Observe that the label accuracy is poorly correlated with strict proof accuracy, and that strict proof accuracy may underestimate the model's reasoning ability.
%\hh{Point out that it may underestimate the label accuracy?}
Rather, the most permissive accuracy metric has the highest correlation with label accuracy, suggesting that label accuracy is appropriate to measure reasoning accuracy. It also suggests that the most permissive proof accuracy metric is most appropriate for measuring the reasoning ability of the model.

\subsection{Proof analysis results} \label{sec:proof_analysis}

%\hh{Can we structure the results by research questions or takeaways (e.g., ontology type, number of hops, model sizes etc.)? Unlike a method paper, we need to guide the reader through the results as it's unclear what they should be looking for.}

%\paragraph{Proof accuracy vs model size.}
\paragraph{Only the largest model is able to reason.}
%\hh{Maybe we should move figure 6 to the appendix, and instead add this paragraph about emergent ability at the beginning. And say that we focus our analysis on instruct-gpt-002 in the rest of the results.}
We investigated how reasoning ability is affected by model size. In figure \ref{fig:smaller_gpt3_results} in the Appendix, proof accuracy increases considerably when increasing the model size from 350M to 1.3B and 6.7B.
%\hh{I thought emerging means that the performance jumps from random (not able to do the task) to much better than random (can do the task) at a certain model/data size.} \abucomment{actually this raises an interesting point: it can be argued that ``chance'' is not 50\% when measuring proof accuracy. a model reaching 50\% means that it has learned to make valid steps but chooses the misleading step at the critical branch point. i'm wondering if we should mention this somewhere. alternative: to avoid this ambiguity, we can move mentioning emergent abilities to after mentioning the result on the 175B 002 model, although that would mean this ability didnt emerge due to scaling-up alone}
However, only \texttt{text-davinci-002} is able to perform better than chance. We were not able to conclusively discern the cause of the significant difference in performance between version \texttt{001} and \texttt{002}. One possible factor is the maximum token limit of version \texttt{002} is roughly twice that of version \texttt{001}. In fact, the model \texttt{davinci} seems to perform as well as, if not slightly better than, \texttt{text-davinci-001}. In addition, we notice that the frequency of invalid steps decreases as the model size increases, and so larger models seem to be better at making valid steps, whether or not those steps are actually useful.

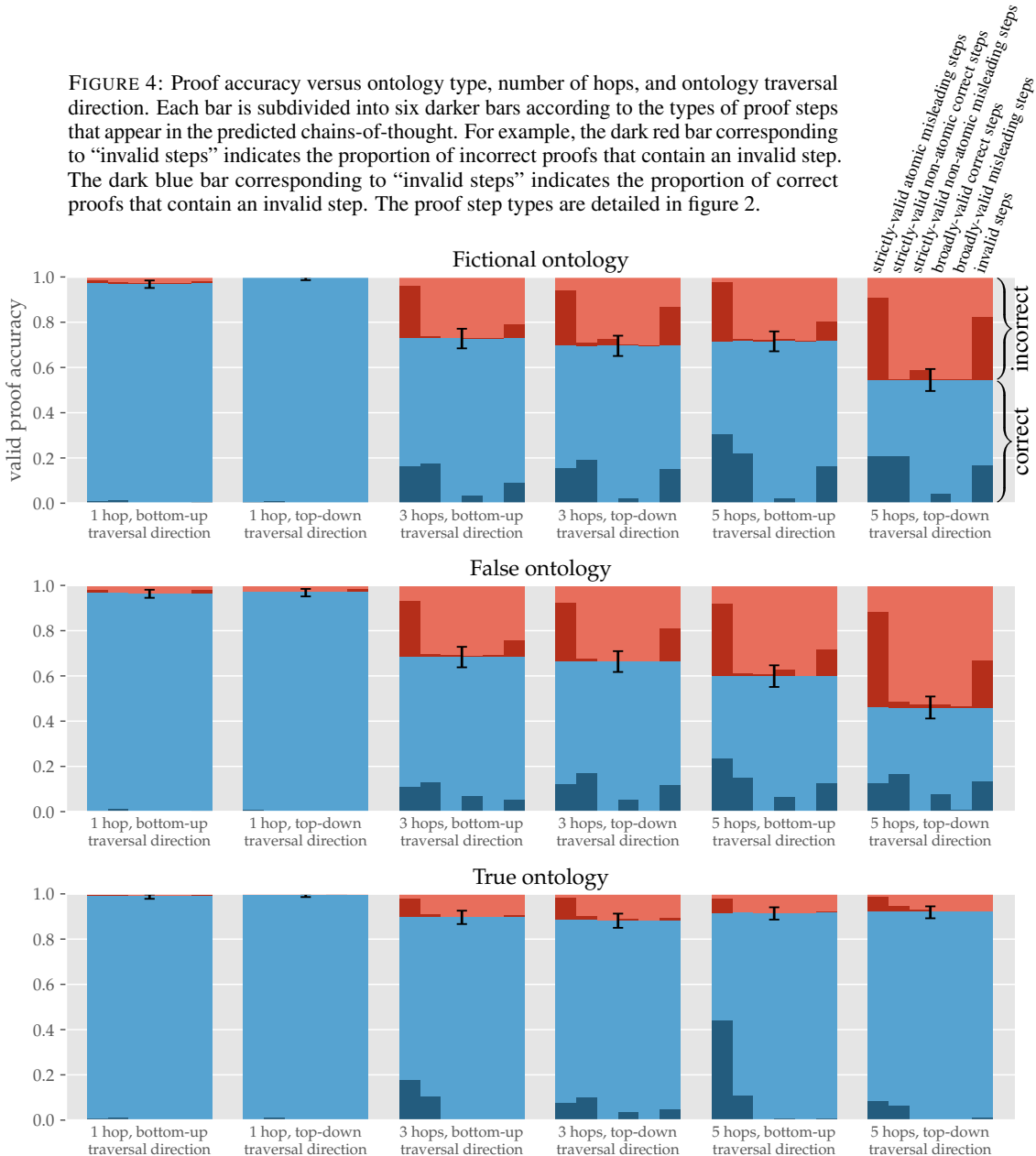
\begin{figure}
    \vspace{-1.0em}
    \floatbox[
            \capbeside
            \thisfloatsetup{capbesideposition={top,left},capbesidewidth=0.8\textwidth}
        ]{figure}[\textwidth]{
        \vspace{-3.2em}
        \hspace{-1.7\textwidth}\makebox[\textwidth][c]{\scalebox{0.7}{\begin{tikzpicture}
            \node[anchor=south west,inner sep=0] (image) at (0,0) {\includegraphics{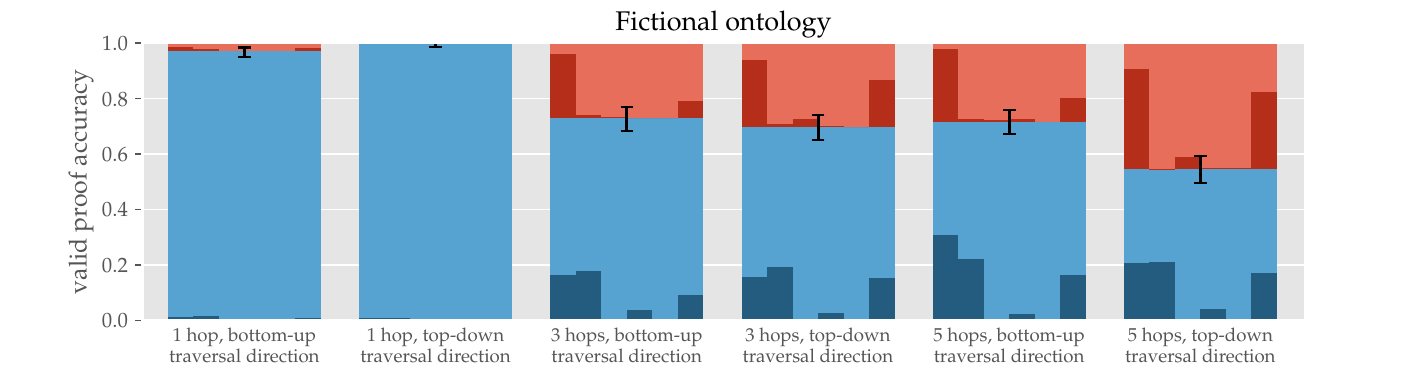}};
            \draw [decorate, decoration={calligraphic brace,amplitude=8pt}, line width=1.4pt] (2.5 + 0.63 + 5*0.76 + 6*2.46 + 0.04, 0.96 + 4.64*0.539) -- (2.5 + 0.63 + 5*0.76 + 6*2.46 + 0.04, 0.96 + 0) node[pos=0.5,rotate=90,yshift=-14pt,black]{\Large correct};
            \draw [decorate, decoration={calligraphic brace,amplitude=8pt}, line width=1.4pt] (2.5 + 0.63 + 5*0.76 + 6*2.46 + 0.04, 0.96 + 4.64*1.0) -- (2.5 + 0.63 + 5*0.76 + 6*2.46 + 0.04, 0.96 + 4.64*0.549) node[pos=0.5,rotate=90,yshift=-13pt,black]{\Large incorrect};
            \node[rotate=70,anchor=south west] at (2.5 + 0.63 + 5*0.76 + 5*2.46 + 0.21 + 0*0.42, 0.96 + 4.64*0.98) {strictly-valid atomic misleading steps};
            \node[rotate=70,anchor=south west] at (2.5 + 0.63 + 5*0.76 + 5*2.46 + 0.21 + 1*0.42, 0.96 + 4.64*0.98) {strictly-valid non-atomic correct steps};
            \node[rotate=70,anchor=south west] at (2.5 + 0.63 + 5*0.76 + 5*2.46 + 0.21 + 2*0.42, 0.96 + 4.64*0.98) {strictly-valid non-atomic misleading steps};
            \node[rotate=70,anchor=south west] at (2.5 + 0.63 + 5*0.76 + 5*2.46 + 0.23 + 3*0.42, 0.96 + 4.64*0.98) {broadly-valid correct steps};
            \node[rotate=70,anchor=south west] at (2.5 + 0.63 + 5*0.76 + 5*2.46 + 0.25 + 4*0.42, 0.96 + 4.64*0.98) {broadly-valid misleading steps};
            \node[rotate=70,anchor=south west] at (2.5 + 0.63 + 5*0.76 + 5*2.46 + 0.27 + 5*0.42, 0.96 + 4.64*0.98) {invalid steps};
        \end{tikzpicture}}} \\
        \hspace{-1.7\textwidth}\makebox[\textwidth][c]{\includegraphics[scale=0.7]{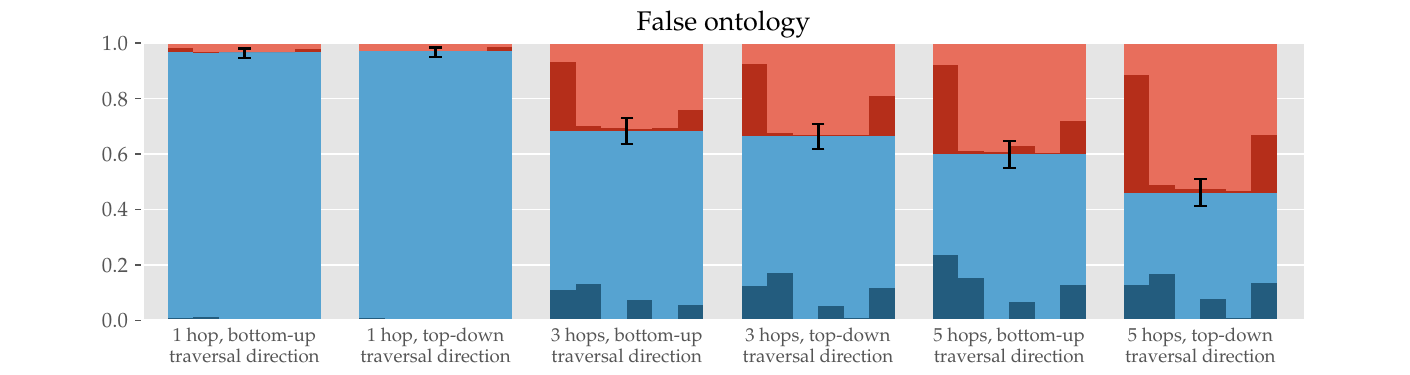}} \\
        \hspace{-1.7\textwidth}\makebox[\textwidth][c]{\includegraphics[scale=0.7]{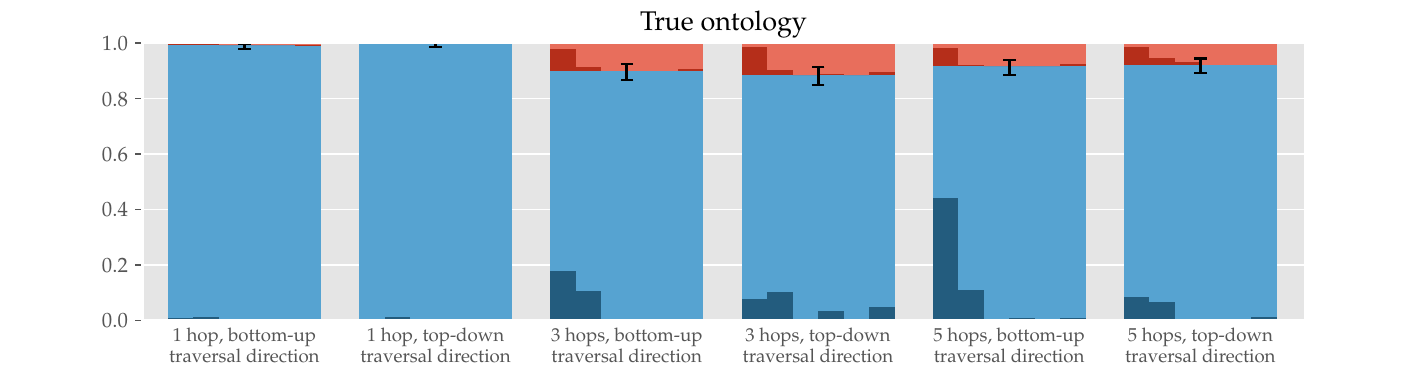}}
    }{\caption{Proof accuracy versus ontology type, number of hops, and ontology traversal direction. Each bar is subdivided into six darker bars according to the types of proof steps that appear in the predicted chains-of-thought. For example, the dark red bar corresponding to ``invalid steps'' indicates the proportion of incorrect proofs that contain an invalid step. The dark blue bar corresponding to ``invalid steps'' indicates the proportion of correct proofs that contain an invalid step. The proof step types are detailed in figure \ref{fig:proof_step_types}.}\label{fig:textdavinci002_proof_accuracy}}
    \vspace{-0.3em}
\end{figure}

For the remainder of the paper, our results focus on \texttt{text-davinci-002}. Our main results are in figure \ref{fig:textdavinci002_proof_accuracy} where we show the proof accuracy and distribution of proof step types for all experiments.
%\hh{Maybe mention that our main results are in figure 3 where we plot the accuracy and distribution of proof step types for all experiments.}

%\paragraph{Proof accuracy vs ontology type.}
\paragraph{Real-world knowledge helps reasoning.}
We investigate the extent to which reasoning ability is affected by whether the ontology is fictional, ``true,'' or ``false.'' Evidently from figure \ref{fig:textdavinci002_proof_accuracy}, the LLM seems to perform comparably in the fictional and ``false'' ontology settings (accuracy is slightly worse with a ``false'' ontology). But when using the ``true'' ontology, the model performs much better, and its performance does not drop when increasing the number of hops from 3 to 5. The model is able to utilize its background knowledge from pretraining to ``skip'' hops, and is thus not as negatively affected by the increased hops. This is consistent with the findings of \citet{DBLP:journals/corr/abs-2207-07051}.

Evidently, the model's reasoning is heavily reliant on real-world knowledge, and this may be a problem for generalizability, such as when applying LLMs to novel scenarios or to settings that are not well-represented in the training data.

%\paragraph{Proof accuracy vs proof length.}
\paragraph{Longer proofs are still challenging.}
We investigate the extent to which reasoning ability is affected by the number of hops in the proof. We see from figure \ref{fig:textdavinci002_proof_accuracy} that the model handles 1- and 3-hop examples quite well but struggles with 5-hop top-down examples, with accuracy falling to chance. So while it is able to perform reasoning to an extent, it is more limited as the number of hops increases.

\paragraph{Traversal direction affects reasoning.}
We also tested how reasoning ability is affected by the traversal direction of the ontology.
We notice in figure \ref{fig:textdavinci002_proof_accuracy} that as the number of hops increases, the model becomes sensitive to the traversal direction of the ontology (top-down vs bottom-up).
%\hh{I don't understand the reasoning here: why does the gold CoT matter? As long as the CoT demo is top-down, the model has training examples for top-down traversal.} \abucomment{the sentence ordering only affects the context; in the CoT, sentences are forced to be bottom-up since conclusions must follow the premises. i changed ``CoT'' to ``proof steps'' to try to avoid this ambiguity, but could also add ``since the conclusions must follow the premises'' to the end of the next sentence}
This may be due to the fact that the order of the gold proof steps mirrors the bottom-up traversal, and is the reverse of the top-down traversal. Thus, the task may be made more difficult for language models if the context sentences are ordered top-down.

%\paragraph{Analysis of proof step types.}
\paragraph{How do LLMs reason step-by-step?}
We investigate the fraction of correct and incorrect proofs that contain various types of proof steps, %\hh{This sentence is a bit misleading as it suggests we are breaking down the steps by types, but actually we are breaking down the proofs by the step types contained in them.}
and whether the correctness of the proof is correlated with the presence of specific types of proof steps.
Figure \ref{fig:textdavinci002_proof_accuracy} breaks down the bars further (in darker red and blue) to indicate the fraction of proofs that contain proof steps other than canonical steps, since most predicted proof steps were canonical (in the 5-hop experiments with fictional ontology, they constitute $93.2\%$ of proof steps).
%\hh{Hmm, but what is the proportion of the canonical steps?}
%\hh{Minor: maybe we should refer to the SVAC steps as canonical steps (to shorten the name).}
%\hh{I wonder if we should talk about the three dimensions separately to make the point crisper (although we'll sacrifice precision) e.g.,
%1) most steps are strictly-valid (in both correct and incorrect ones);
%2) models tend to skip step like humans do (non-atomic steps);
%3) mistakes are due to misleading and invalid steps.
%}
%\abucomment{i edited this as suggested, though im wondering if we should go further and make it into a numbered/bulleted list? but maybe not for the sake of space}\hh{I agree, these are interesting and worth highlight if space allows.}
We make the following observations:
\begin{enumerate}[noitemsep,topsep=-0.3pt,leftmargin=*]
    \item Most predicted proof steps are strictly-valid (in the 5-hop experiments with fictional ontology, $93.2\%$ of proof steps are strictly-valid, $2.4\%$ are broadly-valid, and $5.9\%$ are invalid).
    \item LLMs tend to skip steps by producing non-atomic steps, just as humans do when they verbalize their reasoning (in the 5-hop experiments with fictional ontology, $2.4\%$ of proof steps are non-atomic, even though all steps in the few-shot examples are atomic).
    \item Most incorrect proofs contain misleading steps and invalid steps. This suggests that the source of the incorrect reasoning is either a due to a misleading step or an invalid step that causes the model to produce steps that do not belong to the gold proof.
\end{enumerate}
%We observe that most predicted proof steps are strictly-valid, in both correct and incorrect proofs.
%We also note that LLMs tend to skip steps by producing non-atomic steps, just as humans do when they verbalize their reasoning.
%We observe that many correct proofs contain strictly-valid non-atomic steps.
Intriguingly, some correct proofs also contain misleading steps and invalid steps, which implies that the model is sometimes able to recover from these ``mistakes'' and return to the gold proof. We analyze this behavior in greater detail in section \ref{sec:error_analysis}.

\subsection{What leads to a mistake?} \label{sec:error_analysis}
%\vspace{-0.6em}

%\hh{Same as the previous section: the results should by structured by clear research questions or takeaways.}

\begin{figure}
    \vspace{-1.0em}
    \floatbox[
            \capbeside
            \thisfloatsetup{capbesideposition={top,left},capbesidewidth=0.8\textwidth}
        ]{figure}[\textwidth]{
        \vspace{-3.0em}
        \hspace{-1.72\textwidth}\makebox[\textwidth][c]{\scalebox{0.7}{\begin{tikzpicture}
            \node[anchor=south west,inner sep=0] (image) at (0,0) {\includegraphics{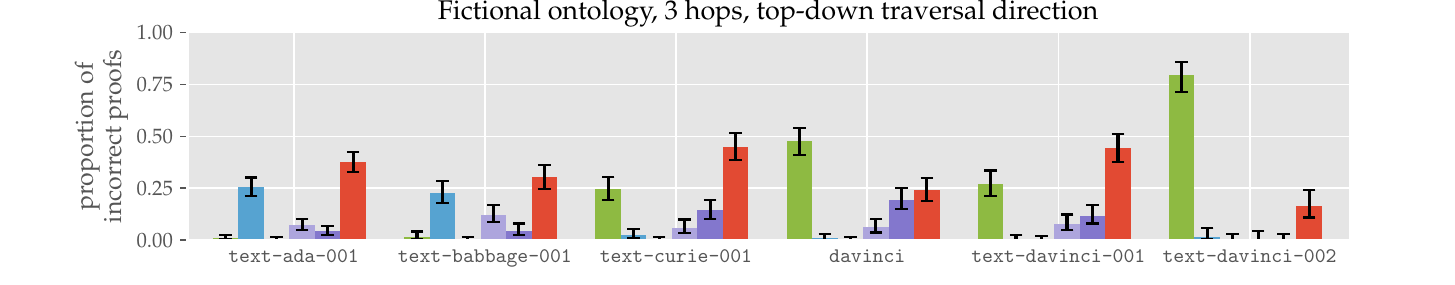}};
            \node[rotate=70,anchor=south west] at (2.5 + 0.63 + 5*0.76 + 5*2.61 + 0.21 + 0*0.42, 0.96 + 3.24*0.98) {strictly-valid atomic misleading steps};
            \node[rotate=70,anchor=south west] at (2.5 + 0.63 + 5*0.76 + 5*2.61 + 0.21 + 1*0.42, 0.96 + 3.24*0.98) {strictly-valid non-atomic correct steps};
            \node[rotate=70,anchor=south west] at (2.5 + 0.63 + 5*0.76 + 5*2.61 + 0.21 + 2*0.42, 0.96 + 3.24*0.98) {strictly-valid non-atomic misleading steps};
            \node[rotate=70,anchor=south west] at (2.5 + 0.63 + 5*0.76 + 5*2.61 + 0.23 + 3*0.42, 0.96 + 3.24*0.98) {broadly-valid correct steps};
            \node[rotate=70,anchor=south west] at (2.5 + 0.63 + 5*0.76 + 5*2.61 + 0.25 + 4*0.42, 0.96 + 3.24*0.98) {broadly-valid misleading steps};
            \node[rotate=70,anchor=south west] at (2.5 + 0.63 + 5*0.76 + 5*2.61 + 0.27 + 5*0.42, 0.96 + 3.24*0.98) {invalid steps};
        \end{tikzpicture}}} \\
        \hspace{-1.72\textwidth}\makebox[\textwidth][c]{\includegraphics[scale=0.7]{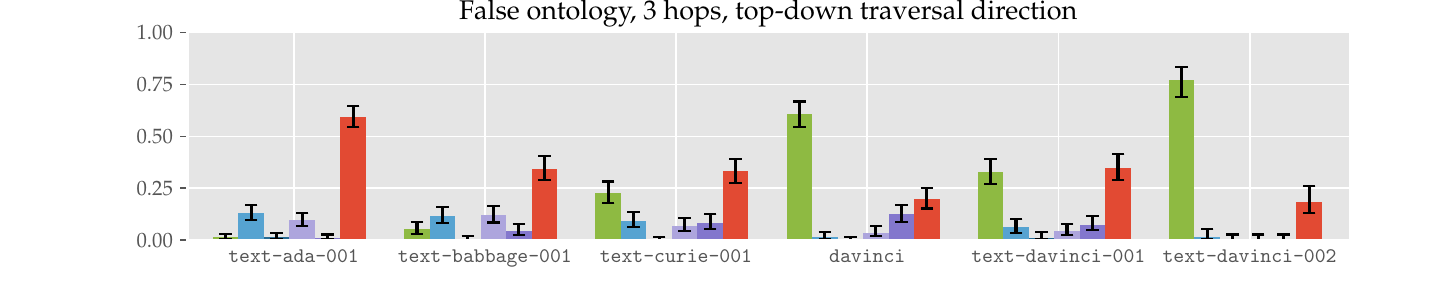}} \\
        \hspace{-1.72\textwidth}\makebox[\textwidth][c]{\includegraphics[scale=0.7]{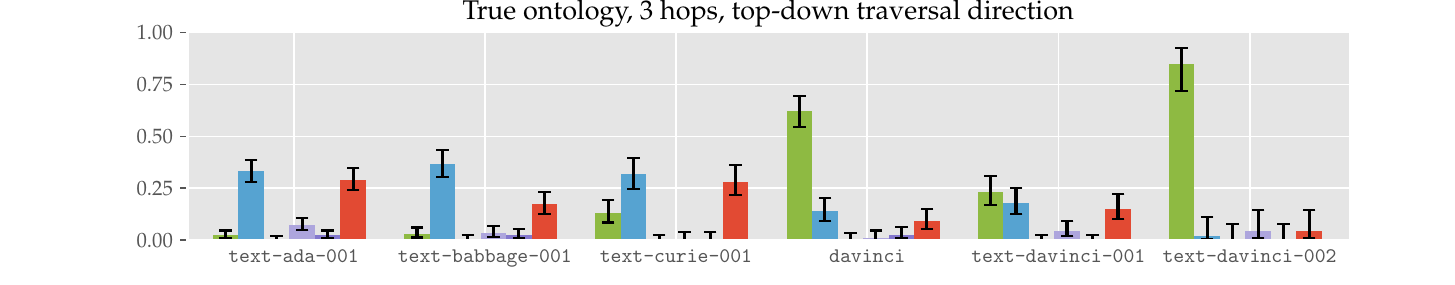}}
    }{\caption{Proportion of incorrect proofs versus the type of the first error (i.e., non-canonical proof step) and model size. The proof step types are detailed in figure \ref{fig:proof_step_types}. We note that in the 3-hop experiments with fictional ontology, four of the 400 examples surpassed the 2049 token limit for all models (except \texttt{text-davinci-002}). These examples were ignored (so the effective number of examples is 396). %\hh{Which examples are ignored?}\abucomment{should we list them in the appendix? they would probably take a good amount of space. or maybe we just list their IDs?}
    We omit the results for the 1-hop experiments here since there were too few incorrect proofs.}\label{fig:first_error}}
    \vspace{2em}
\end{figure}

%\hh{For figure 4, we can probably just keep the last three rows. The changing pattern from small to large models is interesting, but the first three rows are showing the same trend. We can show it in the appendix.}

We investigate whether specific types of proof steps are causing \textsc{InstructGPT} to produce reasoning errors. To do so, we identify the first step in each incorrect proof that is not a canonical step. We observe in figure \ref{fig:first_error}, among incorrect proofs, strictly-valid atomic misleading steps appear in the proof first far more often than other non-canonical step types, including invalid steps. See figure \ref{fig:pronto_example} in the appendix for an example prediction where a misleading step causes the model to fail to prove the goal and produce an invalid step. This indicates that for the best-performing models, the main source of reasoning error is from misleading steps, since most predicted steps are strictly-valid and atomic. That is, imagining the space of proof steps as a graph where each edge represents a single valid step, \textsc{InstructGPT} almost always performs a walk in this graph. Once \textsc{InstructGPT} encounters a branch where one path at the fork follows the correct proof and the other paths do not, \textsc{InstructGPT} will select the incorrect direction with some frequency and is then not able to return to the correct path. Therefore, it seems that while LLMs are able to produce valid proof steps with high probability, they have difficulty with proof planning/strategizing.

We were curious if this relationship held in smaller models. We see in figure \ref{fig:first_error} that smaller models are more prone to make invalid or non-atomic steps as their first non-canonical step. But as model size increases, these types of steps become rarer, and is instead superseded by misleading steps.

Looking again at figure \ref{fig:textdavinci002_proof_accuracy}, we note that many correct proofs also contain misleading steps, and so it must be the case that \textsc{InstructGPT} sometimes returns to the correct proof path at some point after making a misleading step.
%\hh{Maybe we can just report the average steps here and put the entire figure in the appendix, unless you think the distribution conveys additional information.}
To investigate this behavior more closely, we count the number of steps that the model takes \emph{after} making a misleading step until it produces a step in the gold proof and plot the histogram in figure \ref{fig:wrong_branch_lengths} in the appendix. We observe that, in general, the more time the model spends outside the correct proof path, the less likely it becomes to return to the correct proof.

We demonstrate in section \ref{sec:other_prompting} in the appendix that our findings generalize to more sophisticated prompting strategies via an experiment using \emph{self-consistency} prompting \citep{DBLP:journals/corr/abs-2203-11171} and an experiment using a prompt containing example traces of depth-first proof search (i.e. containing examples of the search recovering from misleading steps).

\section{Conclusion and future work}
\vspace{-1.5em}
%\hh{The current version is good, but for improvement, similar to related work, it needs to talk about the role/contribution of different bodies of work in a broader context (reasoning, LLM, human learning etc.), and suggest future directions at a much higher level (e.g., given our results, it could be interesting to study what pretraining data would produce such reasoning ability, or extend it to other dedution rules, or suggesting tasks with long proofs as the next challenge). Currently our future work are immediate next projects :).}
%\hh{The first paragraph can be shortened significantly and described at a higher level than the intro. Let's spend most space here on discussing the implications of the results.}
In this work, we introduced a synthetic fictional QA dataset called \dataset{} designed to evaluate the reasoning ability of LLMs.
%The examples in \abucomment{???} are generated from a fictional ontology, to remove knowledge acquired from pretraining as a confounding factor.
We evaluated \textsc{InstructGPT} and \textsc{GPT-3} on \dataset{} and found that while the largest model was generally able to perform reasoning, it had difficulty with proof planning and selecting the correct proof step when there are multiple available. %it struggled in the setting with 5-hops and top-down sentence ordering. Rather than relying on the model's predicted labels in our evaluation, we semantically parsed the predicted CoT and reconstructed the underlying proof, which we evaluated directly. By analyzing the proof steps, we were able to find that misleading steps were the primary cause of mistakes in the model's reasoning. Thus, while \textsc{InstructGPT} is able to make valid reasoning steps, it has difficulty with proof planning and selecting the correct proof step when there are multiple available.

%\hh{This and the next paragraph are all nice future directions, but it's perhaps a bit too low level. Before that, can we discuss the relevance of our work in broader context, e.g., 1) understanding the capabilities of LLMs (e.g., are they resembling human reasoning capability, connections to cognitive science-y work) 2) building practical reasoning machines (e.g., what does it say about QA in more complex domains or mathematical theorem proving (which many people are investigating using LLMs), neurosymbolic systems etc.)}

\dataset{}, and our high-level approach more broadly, could be used to compare LLM reasoning with that of humans, and to explore which aspects of human reasoning were acquired by LLMs from their pretraining. As our work has shown that LLMs are able to reason to a degree, it is yet unclear where the model acquired this ability. Are there portions of the pretraining data that teach the model to reason? %\abucomment{what kind of CogSci-ish work did you have in mind?}\hh{I was thinking work from brenden lake, tom griffith, and actually also the deepmind work (ishita). But what you wrote is great!}
Our work shows that CoT prompting is not sufficient for more complex reasoning, such as in mathematical domains, since the reasoning tested in this work is a strict subset of that of general mathematical reasoning. Mathematical proofs contain steps with much higher branching factor, where robust proof planning is instrumental. Rather, our results suggest that reasoning systems may benefit from more sophisticated proof planning/search strategies, such as neurosymbolic approaches where part of the reasoning is done over interpretable symbolic structures. \dataset{} can be used to train new reasoning systems, or to pretrain/fine-tune LLMs to improve their reasoning capability.
%\abucomment{not sure if these need citations}

The inability of LLMs to plan ahead in their reasoning might be related to recent work illuminating the theoretical computational limitations of such models \citep{10.1162/tacl_a_00493}.

Since our analysis was limited to modus ponens, proof lengths of at most 5, and semantically simple sentences, it remains to be seen whether LLMs are able to produce longer proofs, or reason with other deduction rules, or over more semantically complex sentences/logical forms.

%Our results suggest that the model's greedily-predicted CoT may not necessarily maximize the likelihood globally, according to the LLM. As such, we can extend the CoT paradigm where after every predicted step in the CoT, we perform beam search decoding to find the top-$k$ most likely values for the next step. We can then perform beam search over the proof steps (i.e., sentences) rather than over the tokens, in order to hopefully find more globally optimal predicted proofs that lead to the correct answer.

%\abucomment{should we add an ethics statement and/or reproducibility statement? (see https://iclr.cc/Conferences/2023/AuthorGuide)}

\subsubsection*{Reproducibility statement}

All our experiments in the main text were run using the OpenAI API on September $9^{\scriptsize th}$, $10^{\scriptsize th}$, and $11^{\scriptsize th}$, 2022. The self-consistency experiment was run on October $29^{\scriptsize th}$ and $30^{\scriptsize th}$, and the DFS experiment was run on November $16^{\scriptsize th}$ (see section \ref{sec:other_prompting}).
For the sake of reproducibility of the analysis, all model outputs, the code for data generation, and the analysis code are freely available with a permissive open-source license at \href{https://github.com/asaparov/prontoqa}{\texttt{github.com/asaparov/prontoqa}}.
The command \texttt{python analyze\_results.py} produces all figures used in this paper.

\subsubsection*{Acknowledgments}

We thank Vishakh Padmakumar, Richard Yuanzhe Pang, Nitish Joshi, Daniel Khashabi, Nicholas Lourie, and Will Merrill for their helpful and insightful discussion. This research was supported by Open Philanthropy, Samsung Advanced Institute of Technology (Next Generation Deep Learning: From Pattern Recognition to AI), AWS AI, and Cisco Research.

\bibliography{iclr2023_conference}
\bibliographystyle{iclr2023_conference}

\appendix

\raggedbottom
\section{Appendix}
\subsection{Deduction rules}

Figure \ref{fig:restricted_proof_calculus} outlines the two deduction rules that we utilize in \dataset{}.

\begin{figure}[h]
    \centering
    \begin{tabular}{>{\centering\arraybackslash}p{0.42\textwidth} >{\centering\arraybackslash}p{0.55\textwidth}}
        \textbf{Deduction rules in general form} & \textbf{Examples} \\[1.5em]
        \footnotesize
        \hspace{1.0em}$\prftree[r]{Hop}{
            \color{RoyalBlue}f(a)
        }{
            \color{RoyalBlue} \forall x(f(x) \to g(x))
        }{\color{RoyalBlue} g(a)}$
        &
        \footnotesize
        \hspace{2.0em}$\prftree[r]{Hop}{
            \color{RoyalBlue} \texttt{cat}(\texttt{fae})
        }{
            \color{RoyalBlue} \forall x(\texttt{cat}(x) \to \texttt{carnivore}(x))
        }{\color{RoyalBlue} \texttt{carnivore}(\texttt{fae})}$ \newline\vspace{-0.6em}\newline
        \textit{i.e., Given that ``Fae is a cat'' and ``All cats are carnivores,'' we conclude that ``Fae is a carnivore.''}
        \\[3.8em]
    
        \footnotesize
        \hspace{1.0em}$\prftree[r]{Ax}{\color{RoyalBlue} A}$
        &
        \footnotesize
        \hspace{6.0em}$\prftree[r]{Ax}{\color{RoyalBlue}\texttt{cat}(\texttt{fae})}$ \newline\vspace{-0.8em}\newline
        \textit{i.e., Assume that ``Fae is a cat'' is an axiom.}
        \\
    \end{tabular}
    \caption{The two deduction rules that constitute the restricted proof calculus in our experiments. All proofs in \dataset{} are composed of instances of only these two rules. Here, $\color{RoyalBlue}A$ is any expression, and $\color{RoyalBlue}f(a)$ is any expression where the variable $\color{RoyalBlue}x$ in $\color{RoyalBlue}f$ is substituted with any term $\color{RoyalBlue}a$ (and similarly for $\color{RoyalBlue}g(a)$).}
    \label{fig:restricted_proof_calculus}
\end{figure}

\subsection{Avoiding shortcuts} \label{sec:avoiding_shortcuts}

When generating examples in \dataset{}, we have to be careful to remove any shortcuts in the question that would allow the model to ``guess'' the answer without reasoning.
In fact, we find that without any distractors, \textsc{InstructGPT} is able to predict the ``true''/``false'' label almost perfectly. %\abucomment{TODO: maybe run this experiment on more samples to get a more precise number}%\hh{Better to mention this result at the beginning of this paragraph as a motivation for the distractor sentence.}
%\hh{Can we first describe the heuristic in this dataset before going into the solution?}
\textsc{InstructGPT} can utilize the heuristic that whether the queried property is mentioned in the context implies whether or not it is true.
For instance, if the example is asking ``Sally is a cat. True or false: Sally is a vertebrate,'' the model can simply look for a string ``Every \uline{\hspace{1.5em}} is (not) a vertebrate,'' regardless of the content in the blank.
Due to the generative process of these examples, this kind of sentence is guaranteed to appear exactly once in the context.
Thus, to ensure that such a heuristic is not informative, we add a distractor sentence by generating a novel concept that is disconnected from the ontology tree, and we assign to this new concept the negation property that is queried by the question.
So in the above example, if the ontology has the rule ``Every mammal is a vertebrate,'' a possible distractor sentence is ``Every insect is not a vertebrate.''
We insert this distractor sentence into a random position in the context.

\subsection{Example \textsc{InstructGPT} misprediction}

\begin{figure}[H]
    \definecolor{color1}{RGB}{219, 122, 118}
    \definecolor{color3}{RGB}{172, 162, 233}
    \definecolor{color4}{RGB}{251, 193, 94}
    \definecolor{color5}{RGB}{142, 186, 66}
    \begin{minipage}{0.94\textwidth}
        \footnotesize
        \texttt{Q: Every vumpus is a numpus. Each vumpus is dull. Dumpuses are vumpuses. Every dumpus is not hot. Every impus is a dumpus. Impuses are brown. Wumpuses are impuses. Wumpuses are opaque. Tumpuses are wumpuses. Tumpuses are small. Every yumpus is hot. Zumpuses are tumpuses. Zumpuses are feisty. Rompuses are zumpuses. Every rompus is not kind. Each jompus is a rompus. Jompuses are sour. Alex is a zumpus. True or false: Alex is hot.} \\
        \texttt{A:} \\

        \textbf{Expected answer:} \texttt{Alex is a zumpus. Zumpuses are tumpuses. Alex is a tumpus. Tumpuses are wumpuses. Alex is a wumpus. Wumpuses are impuses. Alex is an impus. Every impus is a dumpus. Alex is a dumpus. Every dumpus is not hot. Alex is not hot. False} \\

        \setlength{\lineskip}{0pt}
        \textbf{Predicted answer:} \texttt{\hlc[color4]{Alex is a zumpus. Zumpuses are tumpuses. Alex is a tumpus. Tumpuses are wumpuses. Alex is a wumpus. Wumpuses are impuses. Alex is an impus. Every impus is a dumpus. Alex is a dumpus.} \hlc[color3]{Dumpuses are vumpuses.} \hlc[color4]{Alex is a vumpus. Every vumpus is a numpus. Alex is a numpus.} \hlc[color1]{Numpuses are jompuses.} \hlc[color4]{Alex is a jompus. Jompuses are rompuses. Alex is a rompus. Rompuses are zumpuses. Alex is a zumpus. Zumpuses are feisty. Alex is feisty.} True}
    \end{minipage}
    \caption{\setlength{\lineskip}{0pt}An example from \dataset{} (5 hops with fictional ontology and top-down traversal direction) along with the expected and predicted answer from \textsc{InstructGPT}. Note that most of the predicted steps are canonical (highlighted \hlc[color4]{yellow}). The model makes a single misleading step (highlighted \hlc[color3]{purple}) which causes it fail to prove the goal, and to eventually make an invalid step (highlighted \hlc[color1]{red}).}
    \label{fig:pronto_example}
\end{figure}

\subsection{How we evaluate the chain-of-thought}

%Algorithm \ref{alg:evaluate_cot} details the procedure to evaluate the chains-of-thought predicted by LLMs.

\begin{algorithm}[H]
\footnotesize
\let\oldnl\nl% Store \nl in \oldnl
\newcommand{\nonl}{\renewcommand{\nl}{\let\nl\oldnl}}% Remove line number for one line
\SetNlSty{}{\color{RedOrange}\sffamily}{}
\SetAlgoBlockMarkers{}{}
\SetKwProg{Fn}{function}{}{}
\SetKwIF{If}{ElseIf}{Else}{if}{ }{else if}{else }{}
\SetKw{Continue}{continue}
\SetKwFunction{FEvaluateCoT}{\small evaluate\_cot}
\SetKwFunction{FIsProvable}{\small is\_provable}
\SetKwFor{For}{for}{do}{end}
\SetKwProg{uForEach}{for each}{ do}{}
\SetKwProg{Fn}{function}{}{}
\AlgoDisplayBlockMarkers\SetAlgoVlined
\SetAlCapNameFnt{\small}
\SetAlCapFnt{\small}
\SetNoFillComment
\DontPrintSemicolon
\SetInd{0.0em}{0.8em}
    \Fn{\FEvaluateCoT{context sentences $Q_1,\hdots,Q_m$, \newline
        \phantom{\hspace{9.8em}} predicted chain-of-thought sentences $C_1,\hdots,C_n$, \newline
        \phantom{\hspace{9.8em}} gold chain-of-thought sentences $T_1,\hdots,T_r$}}{
        \For(\tcc*[f]{parse the context}){$i \in 1,\hdots,m$}{
            $L^Q_i = $ \texttt{semantic\_parse(}$Q_i$\texttt{)} \;
        }
        \For(\tcc*[f]{parse the gold chain-of-thought}){$i \in 1,\hdots,r$}{
            $L^T_i = $ \texttt{semantic\_parse(}$T_i$\texttt{)} \;
        }
        initialize $S$ as an empty set \;
        \For(\tcc*[f]{parse and evaluate the predicted chain-of-thought}){$i \in 1,\hdots,n$}{
		    $L^C_i = $ \texttt{semantic\_parse(}$C_i$\texttt{)} \;
		    $(P, k) = $ \texttt{is\_provable(}$L^C_i, \{L^Q_1,\hdots,L^Q_m\}, S $\texttt{)} \;
		    \If{$k \ge 0$}{
		        \tcc{if we wish to use a stricter metric for proof accuracy, we can add conditions here (e.g., requiring atomicity by checking $k = 1$)}
		        add $L^C_i$ to $S$ \;
		    }
		    \If{$P \subseteq \{L^T_1,\hdots,L^T_r\}$ and $L^C_i \notin \{L^T_1,\hdots,L^T_r\}$}{
		        \tcc{the premises are in the gold proof but the conclusion is not}
		        mark $L^C_i$ as a misleading step \;
		    }
        }
        \Return{$L^T_r \in S$} \tcc*[r]{the proof is correct if the final conclusion is provable}
	}
    \Fn{\FIsProvable{logical form $\varphi$, set of axioms $A$, previous conclusions $S$}}{
        \uIf{$\varphi \in A$}{
            \Return{$(\{\varphi\},1)$} \tcc*[r]{provable by Ax step (strictly-valid)}
        }\uElseIf{$\varphi \in S$}{
            \Return{$(\{\varphi\},0)$} \tcc*[r]{already proved by previous step}
        }\uElseIf{$\varphi$ has form $g(c)$ or $\neg g(c)$ for any constants $g$ and $c$}{
            \For{$a \in A \cup S$}{
                \If{$a$ has form $\forall x(\psi \to \gamma)$ where $\gamma[x\mapsto c] = \varphi$}{
        		    $(P,k) = $ \texttt{is\_provable(}$\psi[x\mapsto c], A, S$\texttt{)} \;
        		    \If{$k \ge 0$}{
        		        \Return{$(P \cup \{a\}, k + \mathds{1}\{a\in A\})$} \tcc*[r]{provable by Hop step (strictly-valid)}
        		    }
                }
            }
        }\ElseIf{$\varphi$ has form $\forall x(\psi \to \gamma)$}{
            \tcc{note: we precompute this graph}
            let $G$ be the graph where for any axiom in $A$ with form $\forall x(\alpha \to \beta)$, $\alpha$ and $\beta$ are vertices and there is a directed edge from $\alpha$ to $\beta$ \;
            \If{there is a path in $G$ from $\psi$ to $\gamma$}{
                \tcc{provable with additional deduction rules (broadly-valid)}
                \Return{$($ axioms corresponding to path edges $,$ length of path $)$} \;
            }
        }
        \Return{$(\varnothing, -1)$} \tcc*[r]{this step is not provable (i.e., invalid)}
    }
	\caption{Our algorithm for reconstructing and evaluating the proof from the predicted chain-of-thought, and for computing whether each proof step is valid vs invalid, atomic vs non-atomic, misleading vs correct. Here, we use the notation $\varphi[x\mapsto c]$ to denote the substitution of all occurrences of the symbol $x$ with $c$ in the logical form $\varphi$. We use the helper function \texttt{is\_provable} to compute whether a given logical form $\varphi$ is provable from a set of axioms with one or more deduction rules. The function returns a tuple $(P,k)$ where if $k \ge 0$, $\varphi$ is provable in $k$ steps using the premises $P$. Otherwise, $\varphi$ is not provable.}
	\label{alg:evaluate_cot}
\end{algorithm}

\pagebreak
\subsection{Proof accuracy vs model size}

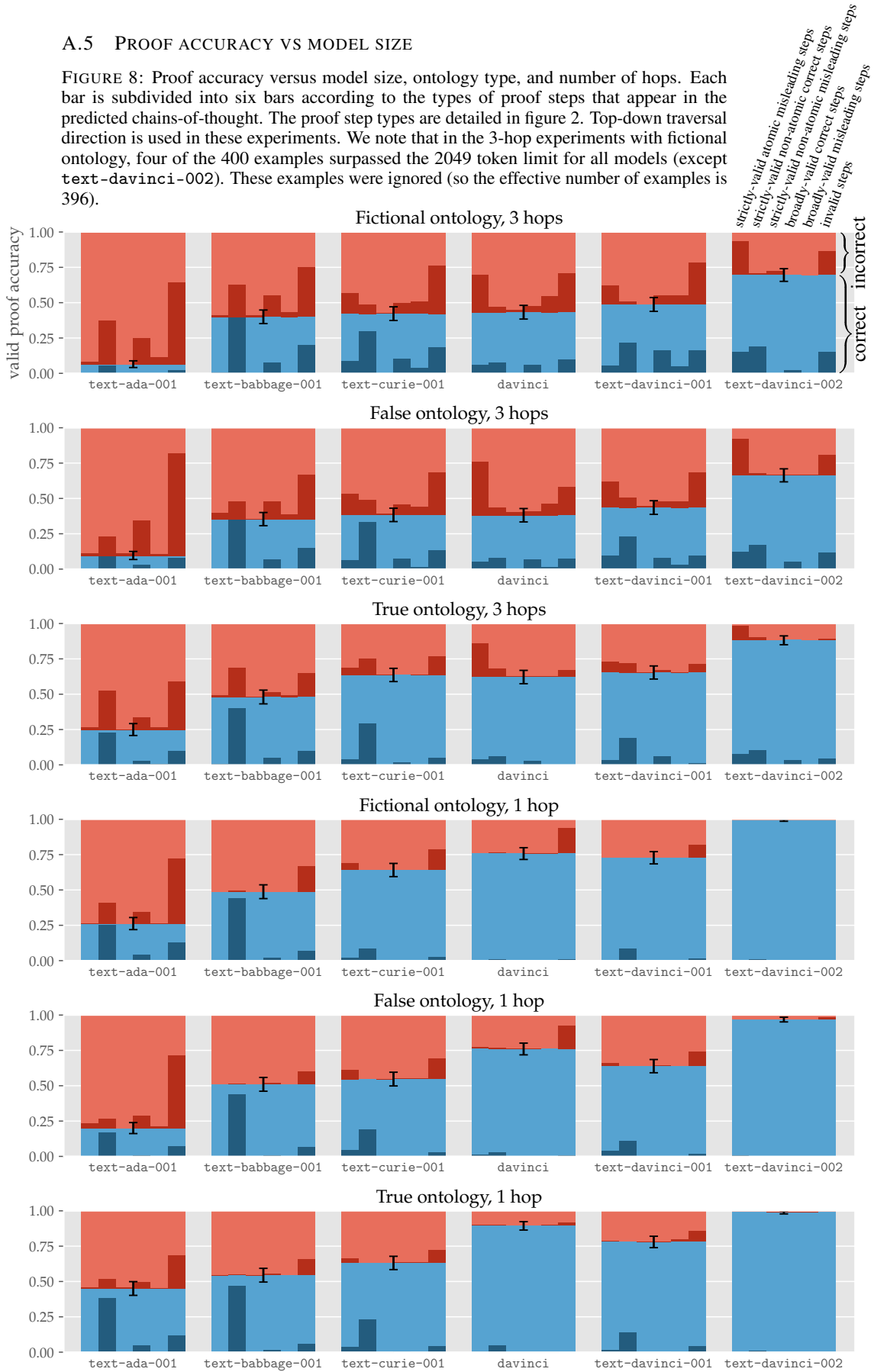
\begin{figure}[H]
    \vspace{-0.9em}
    \floatbox[
            \capbeside
            \thisfloatsetup{capbesideposition={top,left},capbesidewidth=0.83\textwidth}
        ]{figure}[\textwidth]{
        \vspace{-3.6em}
        \hspace{-1.75\textwidth}\makebox[\textwidth][c]{\scalebox{0.7}{\begin{tikzpicture}
            \node[anchor=south west,inner sep=0] (image) at (0,0) {\includegraphics{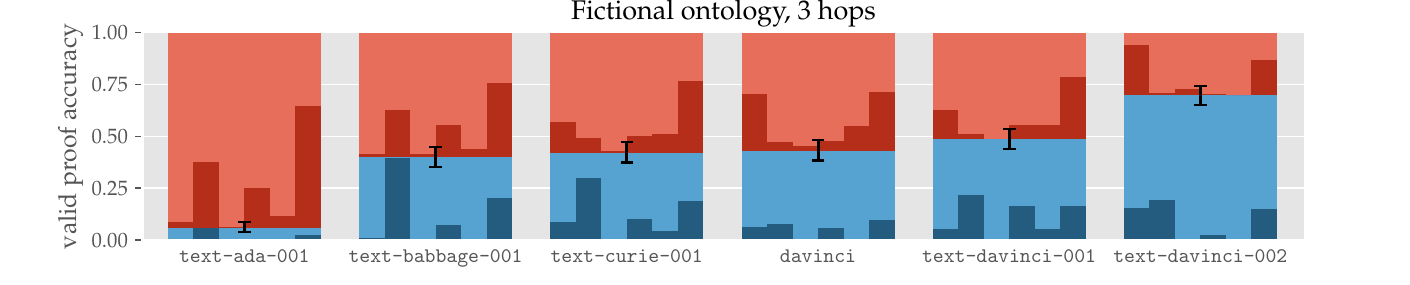}};
            \draw [decorate, decoration={calligraphic brace,amplitude=8pt,aspect=0.6}, line width=1.4pt] (2.5 + 0.63 + 5*0.76 + 6*2.46 + 0.04, 0.78 + 3.48*0.695) -- (2.5 + 0.63 + 5*0.76 + 6*2.46 + 0.04, 0.78 + 0) node[pos=0.6,rotate=90,yshift=-14pt,black]{\Large correct};
            \draw [decorate, decoration={calligraphic brace,amplitude=5pt}, line width=1.4pt] (2.5 + 0.63 + 5*0.76 + 6*2.46 + 0.07, 0.78 + 3.48*1.0) -- (2.5 + 0.63 + 5*0.76 + 6*2.46 + 0.07, 0.78 + 3.48*0.715) node[pos=0.5,rotate=90,yshift=-12pt,black]{\Large incorrect};
            \node[rotate=70,anchor=south west] at (2.5 + 0.63 + 5*0.76 + 5*2.46 + 0.21 + 0*0.42, 0.78 + 3.48*0.98) {strictly-valid atomic misleading steps};
            \node[rotate=70,anchor=south west] at (2.5 + 0.63 + 5*0.76 + 5*2.46 + 0.21 + 1*0.42, 0.78 + 3.48*0.98) {strictly-valid non-atomic correct steps};
            \node[rotate=70,anchor=south west] at (2.5 + 0.63 + 5*0.76 + 5*2.46 + 0.21 + 2*0.42, 0.78 + 3.48*0.98) {strictly-valid non-atomic misleading steps};
            \node[rotate=70,anchor=south west] at (2.5 + 0.63 + 5*0.76 + 5*2.46 + 0.23 + 3*0.42, 0.78 + 3.48*0.98) {broadly-valid correct steps};
            \node[rotate=70,anchor=south west] at (2.5 + 0.63 + 5*0.76 + 5*2.46 + 0.25 + 4*0.42, 0.78 + 3.48*0.98) {broadly-valid misleading steps};
            \node[rotate=70,anchor=south west] at (2.5 + 0.63 + 5*0.76 + 5*2.46 + 0.27 + 5*0.42, 0.78 + 3.48*0.98) {invalid steps};
        \end{tikzpicture}}} \\
        \hspace{-1.75\textwidth}\makebox[\textwidth][c]{\includegraphics[scale=0.7]{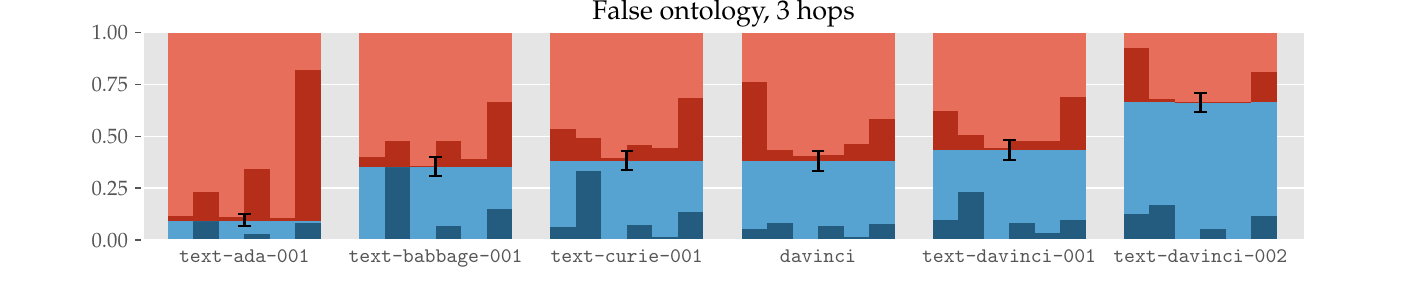}} \\
        \hspace{-1.75\textwidth}\makebox[\textwidth][c]{\includegraphics[scale=0.7]{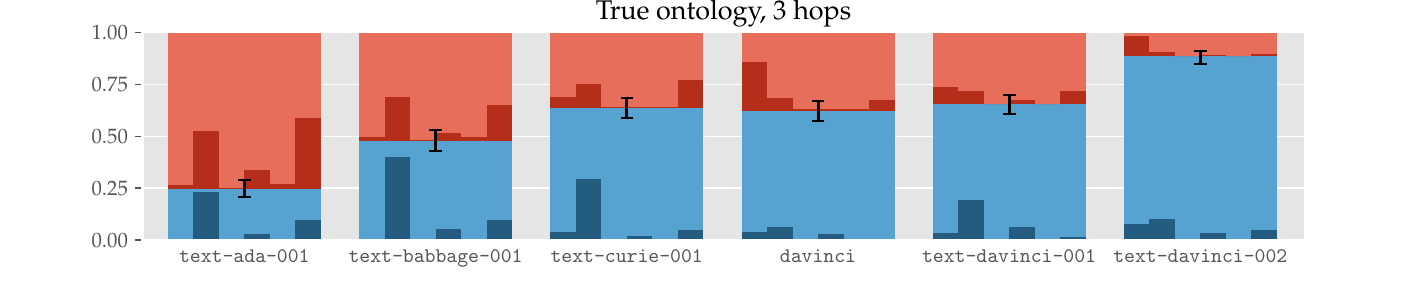}} \\
        \hspace{-1.75\textwidth}\makebox[\textwidth][c]{\includegraphics[scale=0.7]{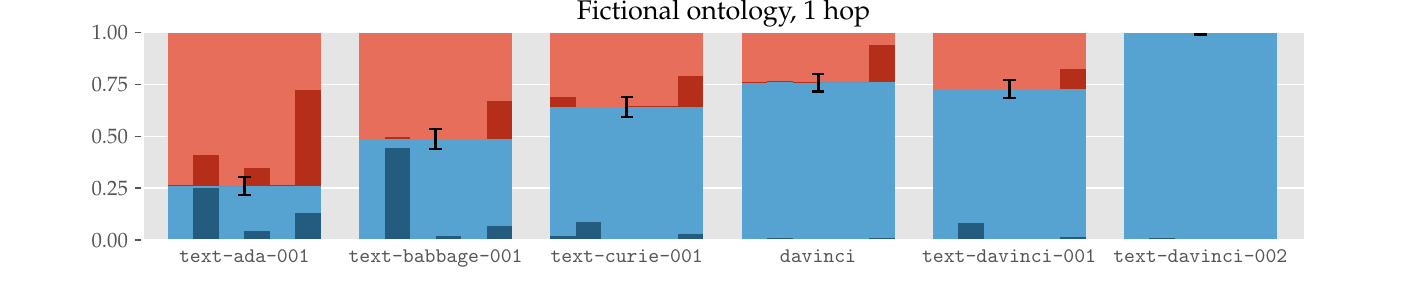}} \\
        \hspace{-1.75\textwidth}\makebox[\textwidth][c]{\includegraphics[scale=0.7]{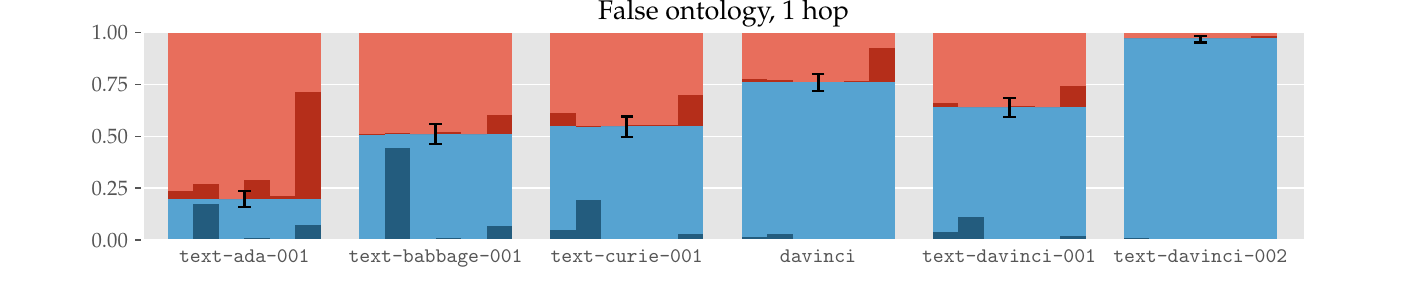}} \\
        \hspace{-1.75\textwidth}\makebox[\textwidth][c]{\includegraphics[scale=0.7]{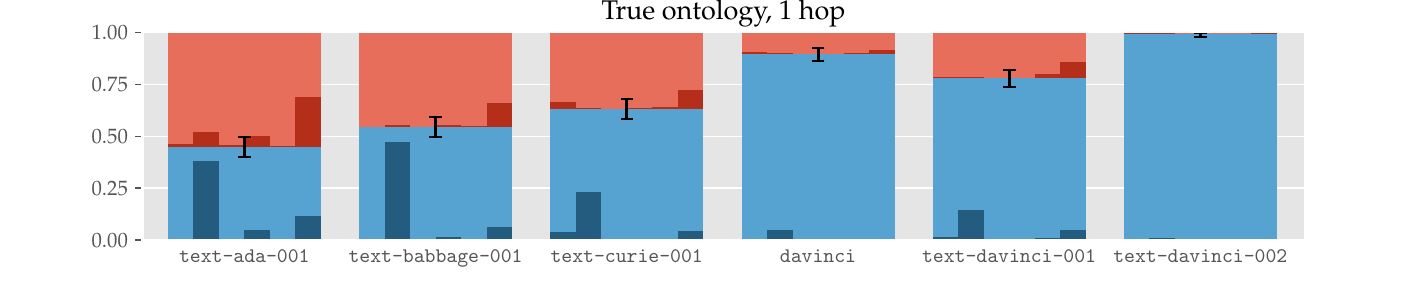}}
    }{\caption{Proof accuracy versus model size, ontology type, and number of hops. Each bar is subdivided into six bars according to the types of proof steps that appear in the predicted chains-of-thought. The proof step types are detailed in figure \ref{fig:proof_step_types}. Top-down traversal direction is used in these experiments. We note that in the 3-hop experiments with fictional ontology, four of the 400 examples surpassed the 2049 token limit for all models (except \texttt{text-davinci-002}). These examples were ignored (so the effective number of examples is 396).}\label{fig:smaller_gpt3_results}}
    \vspace{-4.1em}
\end{figure}

\pagebreak
\subsection{Additional error analysis}

\begin{figure}[H]
    \vspace{-4.0em}
    \floatbox[
            \capbeside
            \thisfloatsetup{capbesideposition={right,center},capbesidewidth=0.25\textwidth}
        ]{figure}[0.75\textwidth]{
        \makebox[0.75\textwidth][c]{\hspace{-1.3em}\scalebox{0.7}{\begin{tikzpicture}
            \node[anchor=south west,inner sep=0] (image) at (0,0) {\includegraphics{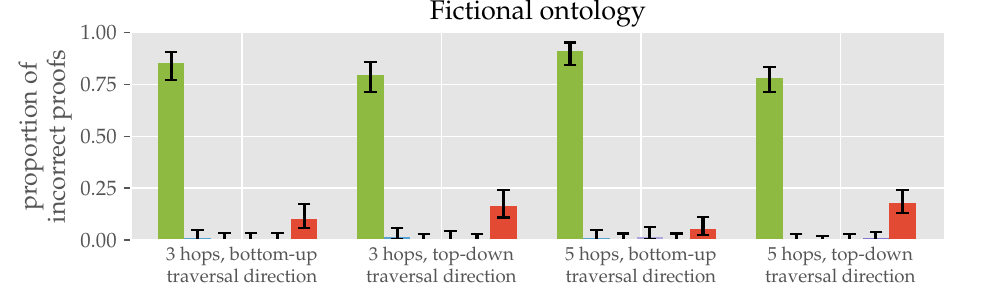}};
            \node[rotate=70,anchor=south west] at (2.5 + 0.63 + 3*0.79 + 3*2.50 + 0.21 + 0*0.44, 0.86 + 3.58*0.98) {strictly-valid atomic misleading steps};
            \node[rotate=70,anchor=south west] at (2.5 + 0.63 + 3*0.79 + 3*2.50 + 0.21 + 1*0.44, 0.86 + 3.58*0.98) {strictly-valid non-atomic correct steps};
            \node[rotate=70,anchor=south west] at (2.5 + 0.63 + 3*0.79 + 3*2.50 + 0.21 + 2*0.44, 0.86 + 3.58*0.98) {strictly-valid non-atomic misleading steps};
            \node[rotate=70,anchor=south west] at (2.5 + 0.63 + 3*0.79 + 3*2.50 + 0.23 + 3*0.44, 0.86 + 3.58*0.98) {broadly-valid correct steps};
            \node[rotate=70,anchor=south west] at (2.5 + 0.63 + 3*0.79 + 3*2.50 + 0.25 + 4*0.44, 0.86 + 3.58*0.98) {broadly-valid misleading steps};
            \node[rotate=70,anchor=south west] at (2.5 + 0.63 + 3*0.79 + 3*2.50 + 0.27 + 5*0.44, 0.86 + 3.58*0.98) {invalid steps};
        \end{tikzpicture}}}
        \makebox[0.75\textwidth][c]{\hspace{-1.3em}\includegraphics[scale=0.7]{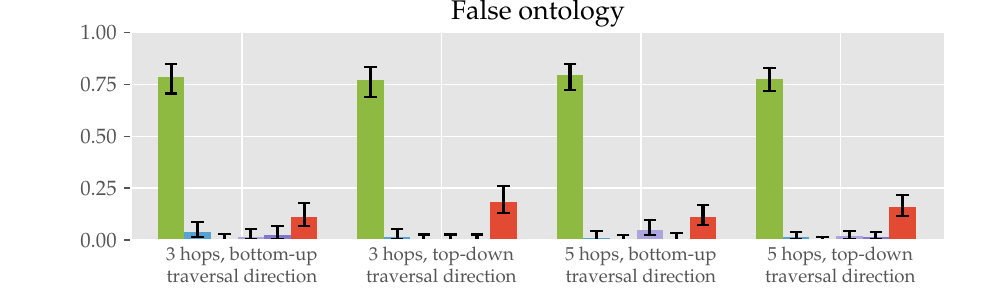}}
        \makebox[0.75\textwidth][c]{\hspace{-1.3em}\includegraphics[scale=0.7]{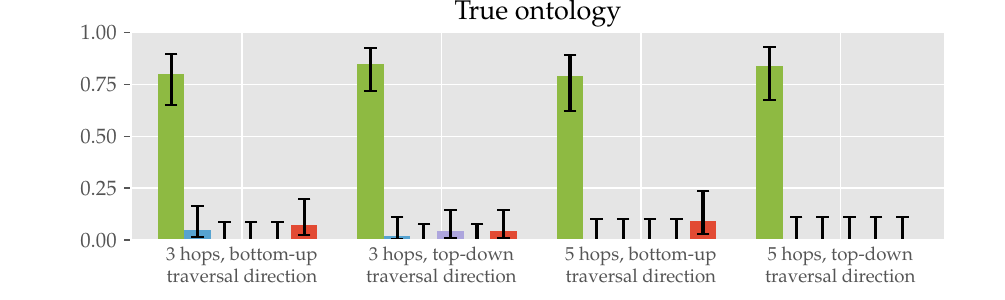}}
    }{\caption{Proportion of incorrect proofs versus the type of the first error (i.e., non-canonical proof step), number of hops, and ontology traversal direction. The proof step types are detailed in figure \ref{fig:proof_step_types}. We note that in the 3-hop experiments with fictional ontology, four of the 400 examples surpassed the 2049 token limit for all models (except \texttt{text-davinci-002}). These examples were ignored (so the effective number of examples is 396). We omit the results for the 1-hop experiments here since there were too few incorrect proofs.}\label{fig:first_error_appendix}}
\end{figure}

\begin{figure}
    \vspace{-3.0em}
    \floatbox[
            \capbeside
            \thisfloatsetup{capbesideposition={right,center},capbesidewidth=0.25\textwidth}
        ]{figure}[0.75\textwidth]{
        \makebox[0.75\textwidth][c]{\hspace{-1.3em}\includegraphics[scale=0.7]{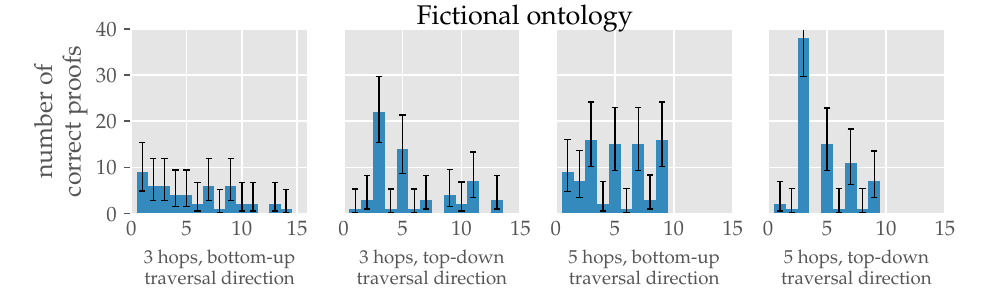}}
        \makebox[0.75\textwidth][c]{\hspace{-1.3em}\includegraphics[scale=0.7]{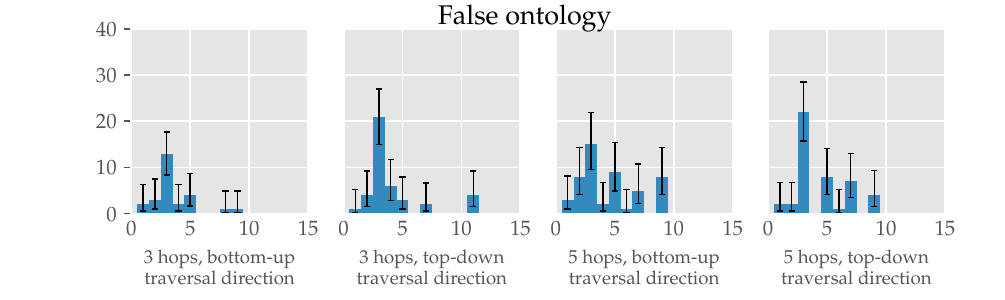}}
        \makebox[0.75\textwidth][c]{\hspace{-1.3em}\includegraphics[scale=0.7]{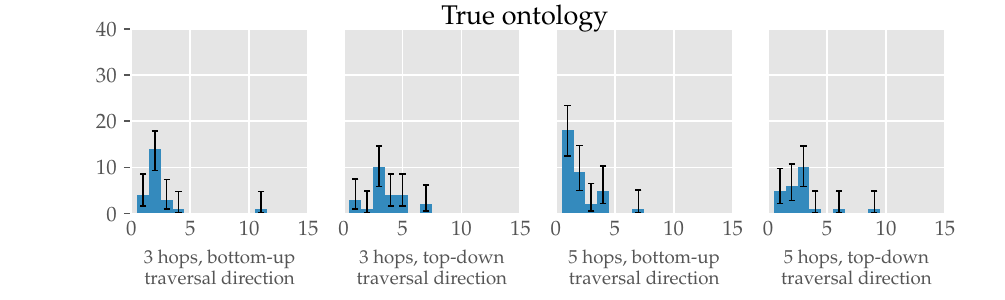}}
    }{\caption{Histograms depicting the distribution of the number of steps in each proof after a strictly-valid atomic misleading step until returning to the gold proof. %\hh{The y-axis is missing "setps"}\abucomment{TODO: do we need all of these? perhaps only keep the text-davinci-002 ones? (i.e. first three rows)}
    }\label{fig:wrong_branch_lengths}}
    \makebox[\textwidth][c]{\includegraphics[scale=0.7]{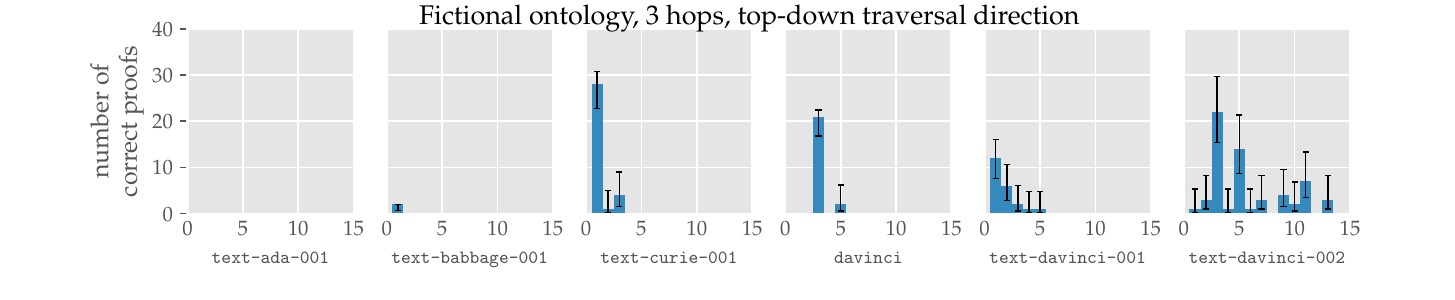}}
    \makebox[\textwidth][c]{\includegraphics[scale=0.7]{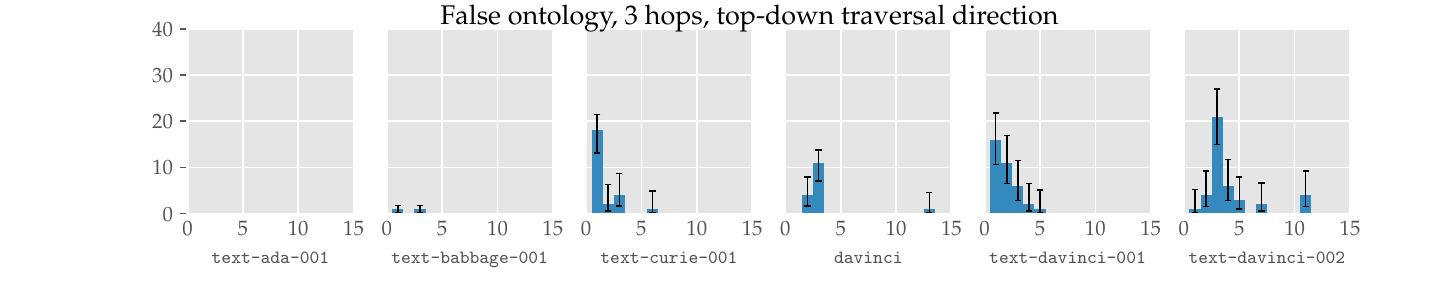}}
    \makebox[\textwidth][c]{\includegraphics[scale=0.7]{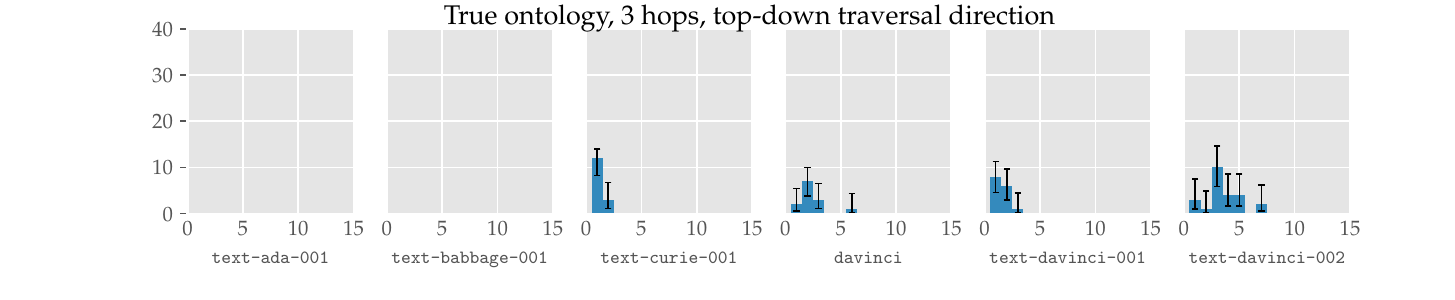}}
    \vspace{-3.0em}
\end{figure}

\pagebreak
\subsection{Do other prompting strategies help?} \label{sec:other_prompting}

\subsubsection{Self-consistency}

To what extent do our findings generalize to prompting strategies other than CoT with greedy decoding? To test this, we experimented with \emph{self-consistency} prompting \citep{DBLP:journals/corr/abs-2203-11171}, where for each example, we queried the LLM for $40$ sample predictions of the CoT, using a temperature setting of $0.7$. For each sample CoT $s_i$, we compute the following quantity:
\begin{equation*}
    \exp\Bigg\{ \frac{1}{|s_i|} \sum_{j=1}^{|s_i|} \log p(s_{i,j} | s_{i,1}, \hdots, s_{i,j-1}) \Bigg\}.
\end{equation*}
We parse each predicted CoT sample into a sequence of logical forms, and we find the logical form sequence with the highest sum of the above quantity over all the CoT samples that share the same semantic parse. This logical form sequence is the final prediction.

We run this experiment in our setting with 5 hops, fictional ontology, and top-down traversal direction, with $100$ examples. The resulting valid proof accuracy is $0.56$ compared to $0.545$ which is not significantly different. Furthermore, inspecting specific examples of CoT samples (see figure \ref{fig:self_consistency_example}), we see that for examples that the model gets wrong, the model is actually assigning higher overall probability to the incorrect proof than to the correct proof. This suggests that our results do in fact generalize to more sophisticated prompting/decoding strategies, and that strategies that endeavor to find proofs with higher probability globally (e.g. beam search) will not help the model in proof planning.

%\abucomment{we could show these results in a plot similar to figure 4 and/or add the results where we aggregate the samples by label (true vs false)}

\begin{figure}
    \definecolor{color1}{RGB}{219, 122, 118}
    \definecolor{color3}{RGB}{172, 162, 233}
    \definecolor{color4}{RGB}{251, 193, 94}
    \definecolor{color5}{RGB}{142, 186, 66}
    \begin{minipage}{0.94\textwidth}
        \footnotesize
        \texttt{Q: Each rompus is a wumpus. Every rompus is not opaque. Every jompus is a rompus. Every jompus is not sour. Vumpuses are jompuses. Each vumpus is not earthy. Yumpuses are vumpuses. Each yumpus is angry. Each tumpus is a yumpus. Tumpuses are large. Numpuses are tumpuses. Numpuses are not cold. Every zumpus is a numpus. Dumpuses are not angry. Zumpuses are shy. Impuses are zumpuses. Impuses are luminous. Polly is an impus. True or false: Polly is angry.} \\
        \texttt{A:} \\

        \textbf{Expected answer:} \texttt{Polly is an impus. Impuses are zumpuses. Polly is a zumpus. Every zumpus is a numpus. Polly is a numpus. Numpuses are tumpuses. Polly is a tumpus. Each tumpus is a yumpus. Polly is a yumpus. Each yumpus is angry. Polly is angry. True} \\

        \setlength{\lineskip}{0pt}
        \textbf{Sample predicted answer:} \texttt{\hlc[color4]{Polly is an impus. Impuses are zumpuses. Polly is a zumpus. Zumpuses are numpuses. Polly is a numpus. Numpuses are tumpuses. Polly is a tumpus. Every tumpus is a yumpus. Polly is a yumpus.} \hlc[color3]{Yumpuses are vumpuses.} \hlc[color4]{Polly is a vumpus. Vumpuses are jompuses. Polly is a jompus. Each jompus is a rompus. Polly is a rompus. Rompuses are wumpuses. Polly is a wumpus.} \hlc[color1]{Wumpuses are not angry.} \hlc[color4]{Polly is not angry.} False}

        \textbf{Sample average log probability:} \texttt{-0.07272354467015028} \\

        $\hdots$ \\

        \textbf{Sample predicted answer:} \texttt{\hlc[color4]{Polly is an impus. Impuses are zumpuses. Polly is a zumpus. Zumpuses are numpuses. Polly is a numpus. Numpuses are tumpuses. Polly is a tumpus. Each tumpus is a yumpus. Polly is a yumpus. Yumpuses are angry. Polly is angry.} True}

        \textbf{Sample average log probability:} \texttt{-0.04729328860082488} \\

        \textbf{Sample predicted answer:} \texttt{\hlc[color4]{Polly is an impus. Impuses are zumpuses. Polly is a zumpus. Zumpuses are numpuses. Polly is a numpus. Numpuses are tumpuses. Polly is a tumpus. Each tumpus is a yumpus. Polly is a yumpus.} \hlc[color3]{Yumpuses are vumpuses.} \hlc[color4]{Polly is a vumpus. Vumpuses are jompuses. Polly is a jompus. Every jompus is a rompus. Polly is a rompus.} \hlc[color1]{Rompuses are not angry.} \hlc[color4]{Polly is not angry.} False}
 
        \textbf{Sample average log probability:} \texttt{-0.02227105943358164} \\

        $\hdots$
    \end{minipage}
    \caption{\setlength{\lineskip}{0pt}An example from \dataset{} (5 hops with fictional ontology and top-down traversal direction) using self-consistency (each sample was produced using a temperature of $0.7$), showing the expected and sample predicted answers from \textsc{InstructGPT}. Canonical steps are highlighted \hlc[color4]{yellow}, misleading steps \hlc[color3]{purple}, and invalid steps \hlc[color1]{red}. We note that the sample predicted CoTs that correspond to the gold proof are given lower overall probability than those that are incorrect.}
    \label{fig:self_consistency_example}
\end{figure}

\subsubsection{Can the model learn to do depth-first search from in-context examples?}

Our earlier analysis also revealed a possible way forward to rectify the model's shortcoming in proof planning: Even after making a misleading step, \textsc{InstructGPT} sometimes ``returned'' to the correct proof. We could instead relax the constraint that the chains-of-thought always reflect the shortest correct proof of the answer. Instead, we allow the in-context examples to contain misleading steps, with the hope that the model learns to better recover from misleading steps. It is reasonable for humans to explore a space of possible solutions before arriving at the correct answer, and we could mimic this in LLMs by allowing the CoT to explore alternative paths in the space of proof steps, even if those paths are not ultimately part of the shortest proof. This kind of search strategy is analogous to depth-first search (DFS) in graphs.

To test whether this approach improves the model's reasoning ability, we conduct an experiment where we provide in-context examples of chains-of-thought that follow a DFS, with the hope that the model is able to learn to perform DFS when given a new test example, thereby improving the likelihood that it finds the correct answer. We run this experiment in our setting with 5 hops, fictional ontology, and top-down traversal direction, with $100$ examples. The resulting valid proof accuracy is $0.55$ compared to $0.545$ which, again, is not significantly different.

\end{document}